\pdfoutput=1

\PassOptionsToPackage{T2A}{fontenc}
\documentclass[multiauthors,onecolumn]{adaption}
\usepackage[table]{xcolor}
\usepackage{soul}
\usepackage{graphicx}
\usepackage{booktabs}
\usepackage{subcaption}
\usepackage{longtable}
\usepackage{pgfplots}
\usepackage{pgfplotstable}
\usepackage{tabularx}
\usepackage{xspace}
\usepackage{graphicx}
\usepackage{multirow}
\usepackage[normalem]{ulem}
\usepackage{pdfpages}
\usepackage{makecell}
\usepackage{fontawesome}
\usepackage{listings}
\usepackage{float}
\usepackage{xcolor}
\usepackage{rotating}
\usepackage{arydshln}
\usepackage{placeins}
\usepackage{amsmath}
\usepackage{footmisc}
\usepackage{tikz}
\usepackage[T1]{fontenc}
\usepackage[utf8]{inputenc}
\usepackage{ltablex}
\keepXColumns
\usepackage{ragged2e}
\usepackage{array}

\usepackage{lmodern}
\usepackage[T1]{fontenc}
\usepackage{xcolor}
\usepackage[most]{tcolorbox}
\usepackage{enumitem}
\usepackage[autostyle,english=american]{csquotes}
\MakeOuterQuote{"}

\definecolor{titleblue}{HTML}{4B5BC4}
\definecolor{sectionbg}{HTML}{D6E0FB}
\definecolor{sectionframe}{HTML}{B9C7EE}
\definecolor{cardbg}{HTML}{F7F9FE}
\definecolor{cardframe}{HTML}{C7D2F0}
\definecolor{phcolor}{HTML}{0B7285}   %
\definecolor{codeblue}{HTML}{2B3A67}

\newcommand{\ph}[1]{{\ttfamily\color{phcolor}\{#1\}}}
\newcommand{\code}[1]{{\ttfamily\color{codeblue}#1}}
 \definecolor{ayadsymbol}{RGB}{76, 110, 230} %
 \definecolor{secondsymbol}{RGB}{140, 165, 240}%
 
 \DeclareSymbolFont{extraup}{U}{zavm}{m}{n}
 \DeclareMathSymbol{\vardiamond}{\mathalpha}{extraup}{87}

\newcommand{\invent}{\textsc{Invent-a-Dataset}\xspace}

\title{Invent a Dataset: Measuring dataset generation abilities with zero seed data}

\author{name={Shivalika Singh}\faa,affiliation={1}}
\author{name={Andrija Djurisic}\faa,affiliation={1}}
\author{name={Gbemileke Onilude},affiliation={1}}
\author{name={Sudip Roy},affiliation={1}}
\author{name={Sara Hooker},affiliation={1}}

\affiliations{
    \item[1] Adaption
}

\graphicspath{{icons/}}
\newcommand{\logo}[1]{\raisebox{-0.15em}{\includegraphics[height=1em]{#1.png}}}

\definecolor{emerald}{rgb}{0.31, 0.78, 0.47}
\definecolor{darkred}{rgb}{0.54, 0, 0}
\definecolor{lightblue}{rgb}{0.66, 0.84, 0.86}
\definecolor{burntorange}{rgb}{0.8, 0.33, 0.0}
\definecolor{awesome}{rgb}{1.0, 0.13, 0.32}
\definecolor{weirdbrown}{rgb}{0.91, 0.58, 0.58}
\definecolor{purple}{rgb}{0.5, 0.0, 0.5}
\definecolor{range}{rgb}{0.1, 0.33, 0.0}
\definecolor{indigo}{rgb}{0.29, 0.0, 0.51}
\definecolor{indigo}{rgb}{0.5, 0.25, 0.75}
\definecolor{goldenrod}{rgb}{0.85, 0.65, 0.13}

\newcommand{\inventapi}{Invent API\xspace}

\pgfplotsset{compat=1.18}

\begin{document}

\begin{abstract}

Building datasets remains one of the most manual and brittle parts of AI development. In this technical report, we focus on the most extreme but also most prevalent setting real world practitioners face: a zero data regime. Here, practitioners don't have any data for the capability they want to learn. We introduce \invent which is a prompt based system to go from dataset description to realistic and large scale post-training datasets. We evaluate \invent against five frontier model APIs including Anthropic~\logo{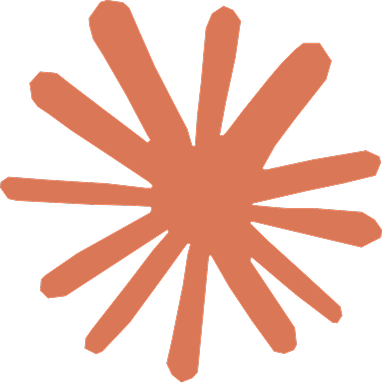}, Google~\logo{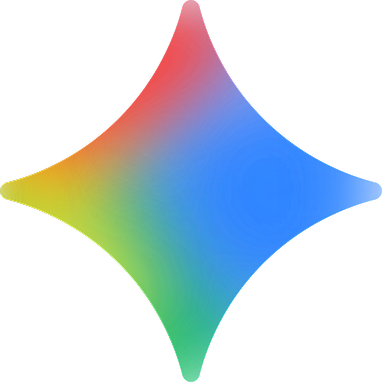}, Open AI~\logo{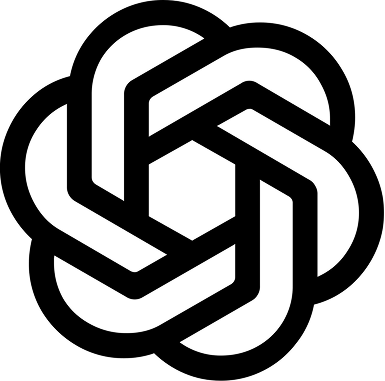}, DeepSeek~\logo{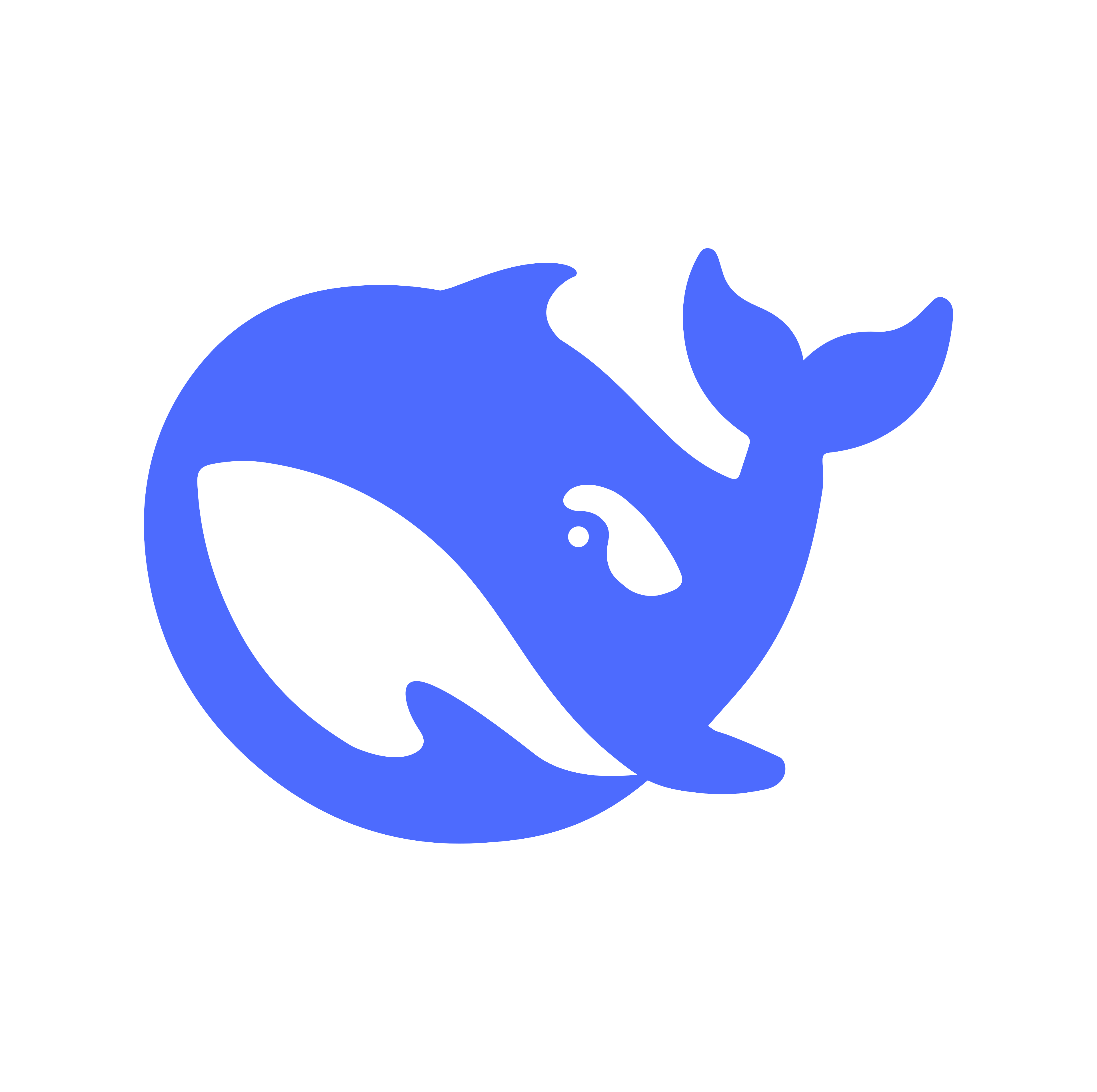}, Zai~\logo{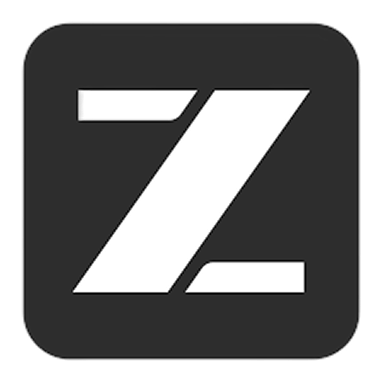}. Across eight task types and dataset sizes up to 20K samples, \invent significantly outperforms with both the highest quality ($17\%$ relative gains) while simultaneously producing the most diverse samples (19\% relative gains). Its diversity advantage widens with scale of training dataset size (from parity at 200 samples to $37\%$ relative gains at 20K samples). This translates into considerable downstream training gains, resulting in far more performant post-trained models. \inventapi fine-tune consistently ranks higher compared to other generator fine-tunes across different post-trained model architectures. 

\small
\textbf{website:} \url{https://adaptionlabs.ai/invent-a-dataset}\\

\normalsize

\end{abstract}

\begin{figure}[h]
    \centering
    \includegraphics[width=0.9\linewidth]{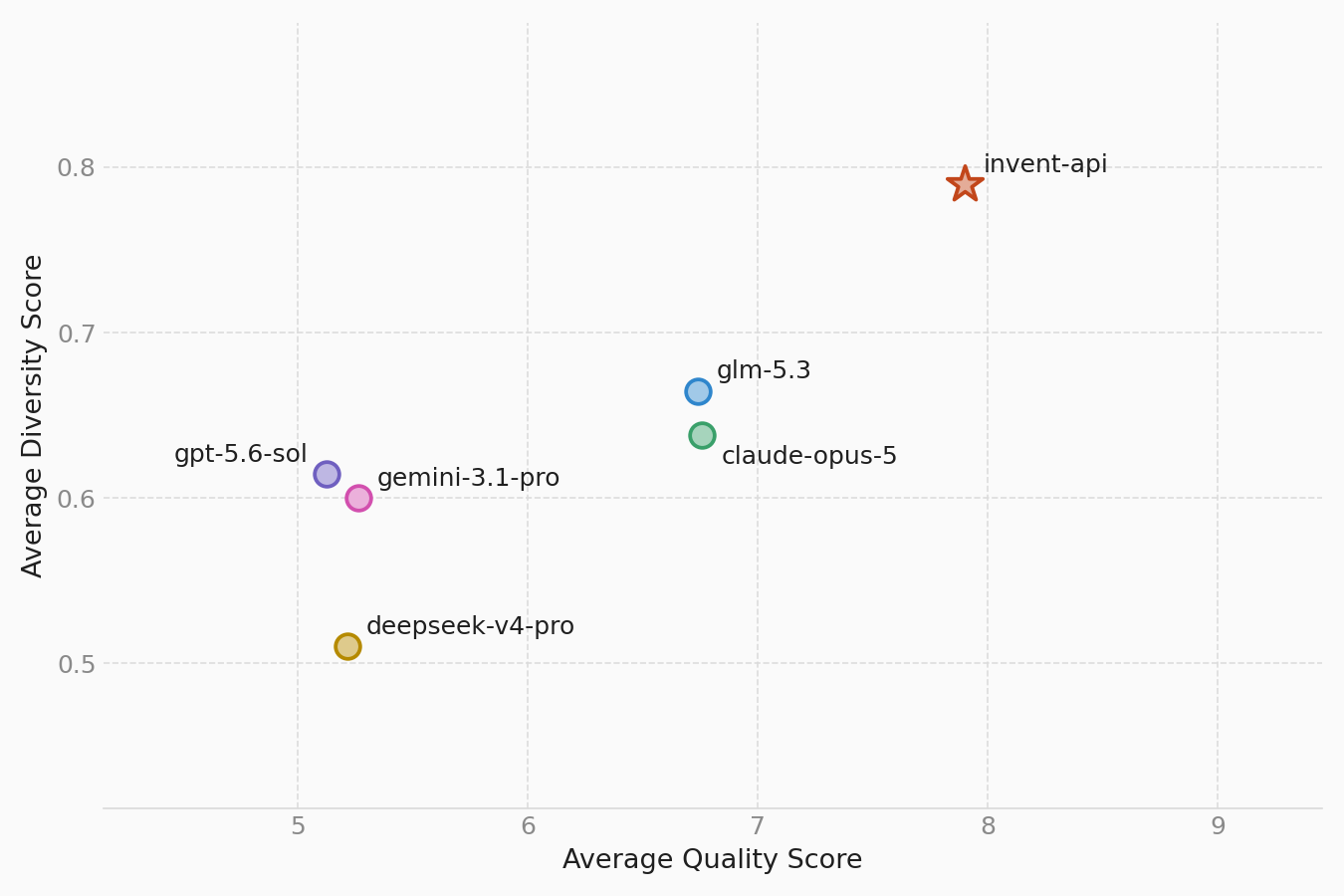}
\caption{\textbf{Diversity vs.\ quality trade-off.} Quality (x-axis) versus diversity (y-axis), measured by DCScore with $\tau = 0.05$, for four external models and the Invent API. Both scores are averaged over eight 5K-sample datasets in unconstrained setting. Higher is better on both axes; the upper right is optimal. The Invent API attains high quality and high diversity simultaneously.}
\label{fig:diversity-quality-5k}
\end{figure}

\section{Introduction}
\label{sec:introduction}
 
Datasets are the raw materials that fuel breakthroughs. Progress at the AI frontier increasingly depends on high-quality training data that continues to challenge models \citep{kulikov2026autodataagenticdatascientist}. Lack of access to data is also one of the biggest determinants of who gets to shape the frontier of AI \citep{longpre2024large}. However, securing high quality data and sufficient volume to target new model capabilities remains incredibly challenging. Human-annotated training datum are expensive to collect and have historically been static representations of the world \citep{kulikov2026autodataagenticdatascientist,salazar2026kaleidoscope,singh2025global,singh-etal-2024-aya,boubdir2023promptsmakedifferencedata}. It takes considerable time to collect additional data, which results in datasets very far from the rich, ever-evolving environment we navigate as humans \citep{roh2019survey}.

In this technical report, we focus on the most extreme end of this data divide. \textbf{A zero training data regime where no data exists for a desired capability.} This is a setting we take seriously because it is also the most common setting real world practitioners face. Several solutions have been proposed for this zero data regime, but fail to guarantee both diversity and quality of the final dataset. Dataset retrieval agents use internet search to comb for relevant datasets but this doesn’t guarantee returning datasets that are formatted correctly for AI training \citep{lin2025rethinking, chapman2020dataset, li2025datasetresearch}. Retrieved datasets are also unlikely to satisfy all the unique constraints a practitioner may have such as formatting, language coverage, output tone. This limits the feasibility of introducing new properties, or explicitly optimizing for task-specific metrics.

Alternatively, practitioners often use existing models to generate synthetic data related to their capability description. However, most synthetic data techniques assume that you either start with a seed pool of prompts or already have a dataset to improve~\citep{wang2023self, xu2024wizardlm}. Without these seed prompts, a significant challenge is scaling to the dataset sizes required for successful post-training while preserving corpus diversity \citep{dohmatob2024model, briesch2023large, shumailov2023curse,bertrand2024stability,guo2024curious}. Large Language Models (LLMs) frequently suffer from low output diversity \citep{holtzman2019curious, xu2206learning}. In the absence of diverse prompts, models with standard maximization-based decoding are known to get stuck in redundant consecutive repetitions. Furthermore, distilling synethtic data for the purpose of training is frequently restricted by licensing: many proprietary state-of-art models have terms of service that explicitly restrict training on their outputs~\citep{wiggers2025openlicenses}. 

In this work, we propose \invent to solve for several of these pain points. \invent enables autonomous, demand-driven data curation that goes from a single dataset description to realistic and AI training ready data ~\citep{adaption2026invent}. We release \invent with permissive licensing that explicitly allows for downstream commercial use in training. We evaluate \invent against directly prompting five frontier model APIs including Anthropic, Google, Open AI, DeepSeek, Zai at diverse dataset sizes up to 20K samples, measuring quality, diversity, and dataset relevance. Our findings can be listed as follows:
\begin{enumerate}
    \item  \inventapi significantly outperforms with both the highest quality ($17\%$ relative gains in quality against the strongest baseline) while simultaneously producing the most diverse samples (19\% at 2k).

\item  This consistent advantage in diversity widens with scale. Unlike most baselines, whose diversity stagnates or declines as datasets requests grow from 2K to 20K samples ($-15\%$ for Claude-Opus-5, $-8\%$ for GLM-5.3, $-7\%$ for GPT-5.6-sol, and $+0.4\%$ for DeepSeek-V4-pro), \inventapi becomes more diverse with scale ($+7\%$) while maintaining stable quality.

\item Adding constraints to the dataset request like tone and output format reduces the diversity of outputs by $15\%$ to $22\%$ for every provider we evaluate. In contrast, \inventapi maintains its diversity for both constrained and the harder unconstrained queries, outperforming the second best, GLM-5.3, by a remarkable $58\%$ at 20K for constrained queries.

\item These quality gains translate to downstream post-training quality. We independently post-train and rank the fine-tuned models' generations across the different data variants. \inventapi fine-tune ranks first on 54\% of the test prompts for Llama-3.3-70B and 41\% for Gemma-4-31B-it, outperforming both the un-tuned base model and the variants fine-tuned on Claude-Opus-5 and GLM-5.3 generated data in each case.  

Our results suggest it is now feasible to automatically target new capabilities from a zero data regime. Historically, the high cost of collecting and curating data has precluded adapting training sets ``on-the-fly'' to increase coverage or task diversity. \invent is an important step toward continuously optimized training sets which can evolve with tasks and target capabilities.

\end{enumerate}

\section{Experimental Setup}

\subsection{Evaluation Setup}

Invent is a dedicated API end-point to generate diverse datasets from a prompt. We benchmark \inventapi against a variety of proprietary and open-weights models. Across each model, desired dataset size and dataset query, we measure the ability to generate a high quality and diverse dataset. We evaluate eight representative dataset queries chosen to represent a range of \textbf{1)} task types, \textbf{2)} languages and \textbf{3)} domains ranging from customer support to medical QA. Additionally, for five of these datasets we compose an even harder version with additional formatting constraints such as tone constraints,json format and multiple choice output restrictions. These are designed to reflect popular practitioner requirements for training datasets \citep{zhou2023instructionfollowingevaluationlargelanguage, jiang2024followbenchmultilevelfinegrainedconstraints, wen2024benchmarkingcomplexinstructionfollowingmultiple}.
We refer to these two sets of evaluation queries as \textbf{unconstrained} and \textbf{constrained} queries respectively. The evaluation queries for three datasets are included in Table~\ref{tab:queries} and remaining in Appendix~\ref{app:dataset-generation-queries}.

\setcounter{table}{1}

\begin{table}[t]
\centering
\footnotesize
\setlength{\extrarowheight}{2pt}
\begin{tabularx}{\textwidth}{@{}%
  >{\RaggedRight\arraybackslash}p{1.7cm}%
  >{\RaggedRight\arraybackslash}X%
  >{\RaggedRight\arraybackslash}X@{}}
\toprule
\textbf{Dataset} & \textbf{Unconstrained Query} & \textbf{Constrained Query} \\
\midrule

African QA &
Dataset containing question-and-answer pairs covering diverse topics related to African research, literature, history, and current events. &
Dataset containing \hl{multi-turn conversations between a teacher and a student on an exam}, covering diverse topics related to African research, literature, history, and current events. \hl{Prompts must use OpenAI format while responses should be plain text.} \\
\midrule

Legal QA &
Dataset containing question--answer pairs focusing on legal advice, factual explanations, and comparative analysis across diverse topics such as copyright law, government types, and liability. &
Dataset containing question--answer pairs focusing on legal advice, factual explanations, and comparative analysis across diverse topics such as copyright law, government types, and liability. \hl{Questions should offer four choices; answers are limited to A, B, C, or D with no additional text.} \\
\midrule

Serbian Car Ads &
A dataset containing a collection of used car advertisements in Serbian, featuring detailed descriptions of vehicle conditions, service histories, and equipment packages. &
A dataset containing a collection of \hl{summaries of} used car advertisements in Serbian, featuring detailed descriptions of vehicle conditions, service histories, and equipment packages. \hl{Each prompt is an ad summary, and each response is a JSON object containing \texttt{vehicle\_description}, \texttt{service\_history}, and \texttt{equipment\_packages}.} \\
\bottomrule
\end{tabularx}
\caption{The single request queries used to generate datasets. Queries are chosen to represent a range
of 1) task types, 2) languages and 3) domains. We evaluate on both \textbf{unconstrained} and \textbf{constrained} versions of the query, where the constrained query requires additional requirements such as 1) tone constraints, 2) json format, 3) multiple choice output restrictions. Differences in the constrained query are \hl{highlighted}. 
}
\label{tab:queries_3}
\end{table}

\textbf{Models Considered.} Our benchmark set includes proprietary models such as Anthropic’s Claude Opus 5 \citep{anthropic2026opus5}, Open AI’s GPT-5.6 Sol \citep{openai2026gpt56}, Google’s Gemini-3.1-pro \citep{google2026gemini31pro}. For the proprietary APIs we choose, it is against their terms of service to use data outputs for training. However, we include these models for an evaluation purpose as an upper bound in performance that is meaningful to calibrate against. Additionally, many practitioners appear to rely on these models for synthetic data generation despite it being prohibited. We also include open-weight models such as Z-ai’s GLM-5.3 \citep{zai2026glm53} and Deepseek AI's DeepSeek-V4-Pro-0813 \citep{deepseek2026v4pro0813} whose license permits use of outputs for downstream training. For our experiments, we use these models via the OpenRouter API \citep{openrouter2026}. 

\begin{figure}[t]
    \centering
    \includegraphics[width=0.8\linewidth]{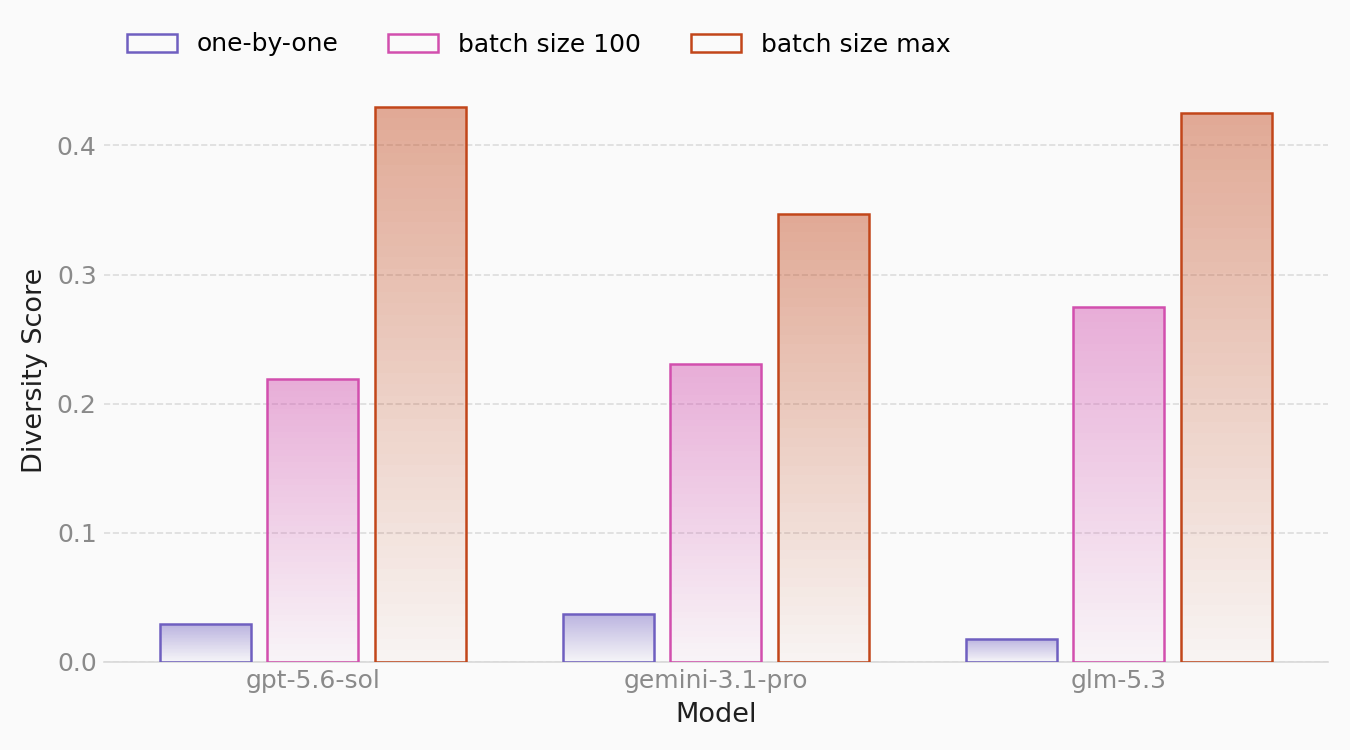}
    \caption{\textbf{Diversity across different generation strategies.} We compare strategies for generating a
    dataset from a single query request (Medical QA) across three models: Gemini 3.1 Pro, GPT-5.6 Sol,
    and GLM-5.3. Strategies differ in how many samples are produced per request: one sample
    at a time (\textit{one-by-one}), fixed batches of 100, or as many samples as fit within the model's
    maximum output-token limit (\textit{batch\_size max}). The y-axis shows DCScore ($\tau = 0.1$),
    where higher is better. Diversity generally increases with batch size, and generating samples one by one
    leads to near-complete diversity collapse ($\leq 0.03$) for every model.}
    \label{fig:generation-strategy}
\end{figure}

\textbf{Generation configurations.} We kept all model hyperparameters at their default values (temperature 1.0 where possible to specify, reasoning disabled). \invent returns a dataset in a single call, however in practice we observed that other models would rarely return the required number of samples if only a single call was used. At scale, single call requests often led to significant diversity collapse.  While we continued to benchmark \invent using only a single call, we evaluated a set of different approaches on the other models benchmark : (i) \textit{one-by-one}, prompting the model to produce a single sample per request (in contrast to requesting entire dataset in one query) (ii) \textit{batch of 100}, producing samples in fixed batches of N=100 (iii) \textit{max batch size}, producing as many samples as possible in a single request depending on model's output token limit which was set to 128K across all models. As shown in Figure~\ref{fig:generation-strategy}, diversity collapses almost entirely ($\approx$0.00--0.03 across all four models) for the \textit{one-by-one} approach, while \textit{max batch} strategy is consistently the most diverse. Based on these results, we selected \textit{max batch} as the default generation approach. We adopt this for all subsequent experiments, unless otherwise specified. 

\textbf{Dataset scaling.} In practice, a core goal of obtaining sufficient data is to train a model. Hence, we are interested in the ability to generate data at a sufficient scale that is viable for training. For each query and model generator we consider, we invent data at four distinct scales: 200, 2000, 5000, 20000 samples. Larger scales allows us to measure how quality and diversity is impacted as the practitioner requests larger datasets for post-training purposes.

\textbf{Downstream post-training set-up.} To test whether using \inventapi translates into downstream model performance gains, we repeat the same procedure independently for three models (Llama-3.3-70B-Instruct, Gemma-4-31B-it and Qwen3.5-9B) corresponding to different model families and model sizes. We post-train each of them on the 20K \textbf{Medical QA} dataset variants produced by every generator (samples shown in Table~\ref{tab:medqa_by_model}) with LoRA SFT.
Rather than hand-tuning a training recipe, we use Autoscientist~\citep{adaption2026autoscientist} to run a hyperparameter search on one of the \textbf{Medical QA} data variants. %
To control for the impact of dataset alone, the configuration proposed by Autoscientist is then applied to every generator's dataset. As a result, every model in the comparison starts from the same base and is trained with the same optimizer settings, for the same number of epochs, on the same number of samples, so the training data is the only variable that differs between them. %

\begin{table}[h!]
\centering
\footnotesize
\setlength{\extrarowheight}{2pt}
\begin{tabularx}{\textwidth}{@{}%
  >{\RaggedRight\arraybackslash}p{3cm}%
  >{\RaggedRight\arraybackslash}p{5cm}%
  >{\RaggedRight\arraybackslash}X@{}}
\toprule
\textbf{API} & \textbf{Prompt} & \textbf{Completion} \\
\midrule

\texttt{Claude Opus 5} &
``How is hypertension diagnosed, and what blood pressure numbers matter?'' &
``Hypertension is diagnosed from properly measured, repeated blood pressure readings rather than a single office value. The patient should be seated quietly for five minutes, back supported, feet flat, \,[\ldots]'' \\
\midrule

\texttt{Gemini 3.1 Pro} &
``What class of anticoagulant is Dabigatran?'' &
``Dabigatran is a Direct Thrombin Inhibitor (DTI). It reversibly binds to the active site of thrombin (Factor IIa), preventing the conversion of fibrinogen to fibrin. It was the first DOAC approved as an  \,[\ldots]'' \\
\midrule

\texttt{DeepSeek V4 Pro} &
``What is the mechanism of action of digoxin?'' &
``Digoxin inhibits the ATPase pump in cardiac myocytes, raising intracellular sodium and, via the sodium-calcium exchanger, intracellular calcium \,[\ldots]'' \\
\midrule

\texttt{GPT 5.6 Sol} &
``What are warning signs of a heart attack?'' &
``Warning signs include pressure, squeezing, heaviness, or pain in the chest, sometimes spreading to an arm, shoulder, jaw, back, or upper abdomen. Shortness of breath, sweating, nausea, dizziness, or unusual fatigue may also occur, and symptoms can be atypical in women, older adults, and people with diabetes.\,[\ldots]'' \\
\midrule

\texttt{GLM 5.3} &
``What are the diagnostic criteria for hypertension?'' &
``Hypertension is diagnosed based on repeated elevated office blood pressure measurements. Under the 2017 ACC/AHA guidelines, stage 1 is systolic 130--139 or diastolic 80--89 mmHg, and stage 2 is 140/90 mmHg or higher. Out-of-office ambulatory or home monitoring is recommended to confirm the diagnosis.\,[\ldots]'' \\
\midrule

\texttt{Invent API} &
``A 62-year-old male presents with progressive shortness of breath, bilateral inspiratory crackles, and a dry cough. He has a 30-year history of asbestos exposure. Discuss the differential diagnosis between idiopathic pulmonary fibrosis and asbestosis.'' &
``\textbf{Differential Diagnosis: IPF vs.\ Asbestosis.} The presentation of progressive dyspnea, dry cough and bilateral `Velcro' crackles with a 30-year asbestos history is a classic diagnostic challenge. Both are chronic, progressive fibrosing interstitial lung diseases sharing a Usual Interstitial Pneumonia (UIP) pattern.\,[\ldots]'' \\
\bottomrule
\end{tabularx}
\caption{Example prompt--completion pairs generated by each API for the \textbf{Medical QA} dataset, whose query was: \textit{``Dataset of medical question-answer pairs covering diagnoses, treatments, drug explanations, and healthcare roles. Each entry includes a prompt and a detailed, expert-generated response''}}
\label{tab:medqa_by_model}
\end{table}

\subsection{Evaluation Metrics}
\label{sec:eval-metrics}
\textbf{Diversity Metrics.} We measure the generated dataset diversity along two complementary axes i.e. \emph{semantic}  and \emph{lexical}.
\begin{itemize}
\item \textbf{Semantic diversity.} As our primary, semantically grounded measure we use the DCScore \citep{zhu2025measuring}. The DCScore is an embedding based metric that treats diversity evaluation as a sample classification task, enabling the metric to capture mutual relationships among the samples.  We use  the Qwen-3-4B-embedding model \citep{zhang2025qwen3} as our embedding model to compute diversity scores since it supports multiple languages and 32K context length which is essential to correctly evaluate the datasets in our benchmark suite.
\item \textbf{Lexical diversity.} In addition to DCScore, we also report metrics related to textual form rather than semantic content computed on the generated \emph{prompts}:
(i) \emph{$n$-gram diversity} (NGD), the summed ratio of unique to total $n$-grams for
$n\in\{1,\dots,4\}$ over the corpus \citep{shaib2026standardizingmeasurementtextdiversity, padmakumar2024doeswritinglanguagemodels}, where higher
is more diverse; (ii) \emph{compression ratio} (CR), the gzip ratio of raw to compressed corpus
size \citep{shaib2026standardizingmeasurementtextdiversity}, where higher indicates more redundancy; and
(iii) \emph{exact-duplicate rate}, the fraction of prompts that are verbatim repeats. 
\end{itemize}

For all these metrics, we report them on a fixed $2{,}000$-prompt random sample from each generated corpus when comparing across dataset scales .

\textbf{Quality Metrics.} We adopt a rubric-based LLM-as-a-Judge methodology to evaluate quality, as it has been to shown by several works to be correlated with human judgement \citep{zheng2023judging, liu2303g}. 
\begin{itemize}
\item \textbf{Prompt and completion quality} is scored on a scale of $0$--$10$. 
Detailed description of these metrics and full judging prompt (see ~\ref{app:quality-judge-template}) along with per-dataset results at various dataset size scales (see ~\ref{app:quality-per-dataset}) have been reported in the Appendix.  
\item \textbf{Relevance} which measures how well generated samples address the underlying intent and scope of the user's request, also scored on the scale from $0$ to $10$. Relevance score validates that a sample matches the request's task type and subject matter, is written in the target language and follows the constraints the request explicitly states (e.g. length, format, structure, style).
\item \textbf{Post-training quality} We employ LLM-as-a-Judge ranking to compare models fine-tuned on Medical QA data from each generator. These are evaluated on a set of in-house, held-out set of medical domain prompts. We don't derive our test set from any single generator's data since such a split would be drawn from that generator's own distribution and would be biased towards that model ~\citep{torralba2011unbiased, NEURIPS2020_045117b0}. Our test prompts are instead sampled from the real-world distribution of medical queries and are held out from every generator's training data. This is a harder, out-of-distribution test and a more faithful indicator of downstream generalization~\citep{pmlr-v97-recht19a, pmlr-v139-koh21a}. For each test prompt, the judge (Gemini 3.1 Pro) sees all responses in a single call and returns a best-to-worst ordering. To control for position bias, the order of the responses is shuffled independently for each prompt, so no model occupies a fixed slot. We provide the judge prompt used for ranking evaluation for reference in Appendix~\ref{app:ranking_evaluation}.
\end{itemize}

\section{Results}
\label{sec:overall_results}

\subsection{Overall Performance: Diversity vs Quality}
\label{sec:overall_performance}

\textbf{Diversity vs Quality Comparison} Figure~\ref{fig:diversity-quality-5k} plots average quality against average diversity (DCScore) at the $5{,}000$-sample scale, averaged over all eight datasets. A strong synthetic-data generator should occupy the upper-right region showcasing it is both diverse \emph{and} high-quality. The \inventapi is the only generator that does so, standing well clear of every external system evaluated on both axes simultaneously. It achieves the highest diversity by a wide margin, roughly $19\%$ above the next-best model (GLM-5.3), $24\%$ above Claude Opus~5, $29\%$ above GPT-5.6 Sol, $32\%$ above Gemini-3.1-pro, and $55\%$ above DeepSeek-V4-Pro. At the same time it also achieves the highest quality score ($\approx 7.9$), about $17\%$ above the strongest baselines (Claude Opus~5 and GLM-5.3, both $\approx 6.75$). Appendix~\ref{app:diversity_all_plots} reports the diversity--quality trade-off at all four dataset scales.

\begin{figure*}[t]
\centering
    \begin{subfigure}[t]{0.45\textwidth} 
    \centering
        \includegraphics[width=\textwidth]{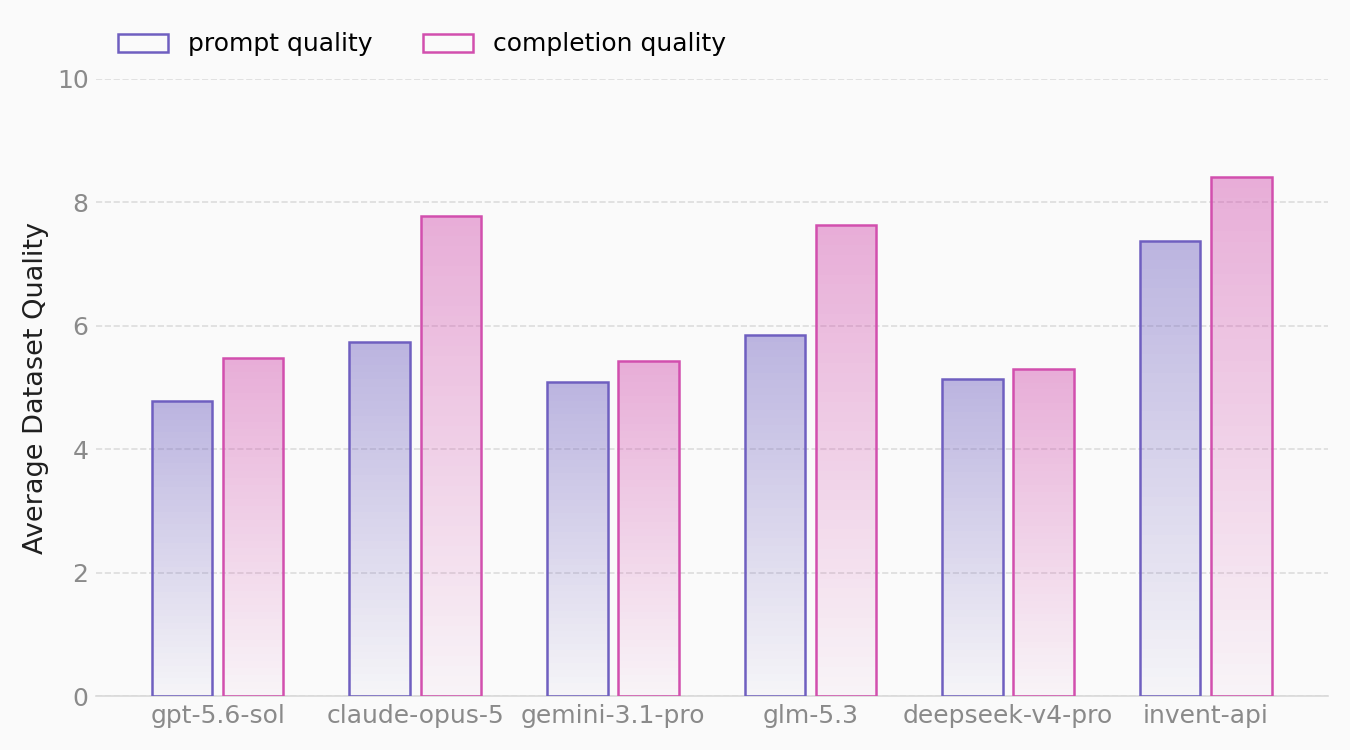}
        \caption{\textbf{Quality per generator}}
    \label{fig:quality-per-model}
    \end{subfigure}
    \begin{subfigure}[t]{0.45\textwidth} 
    \centering
        \includegraphics[width=\textwidth]{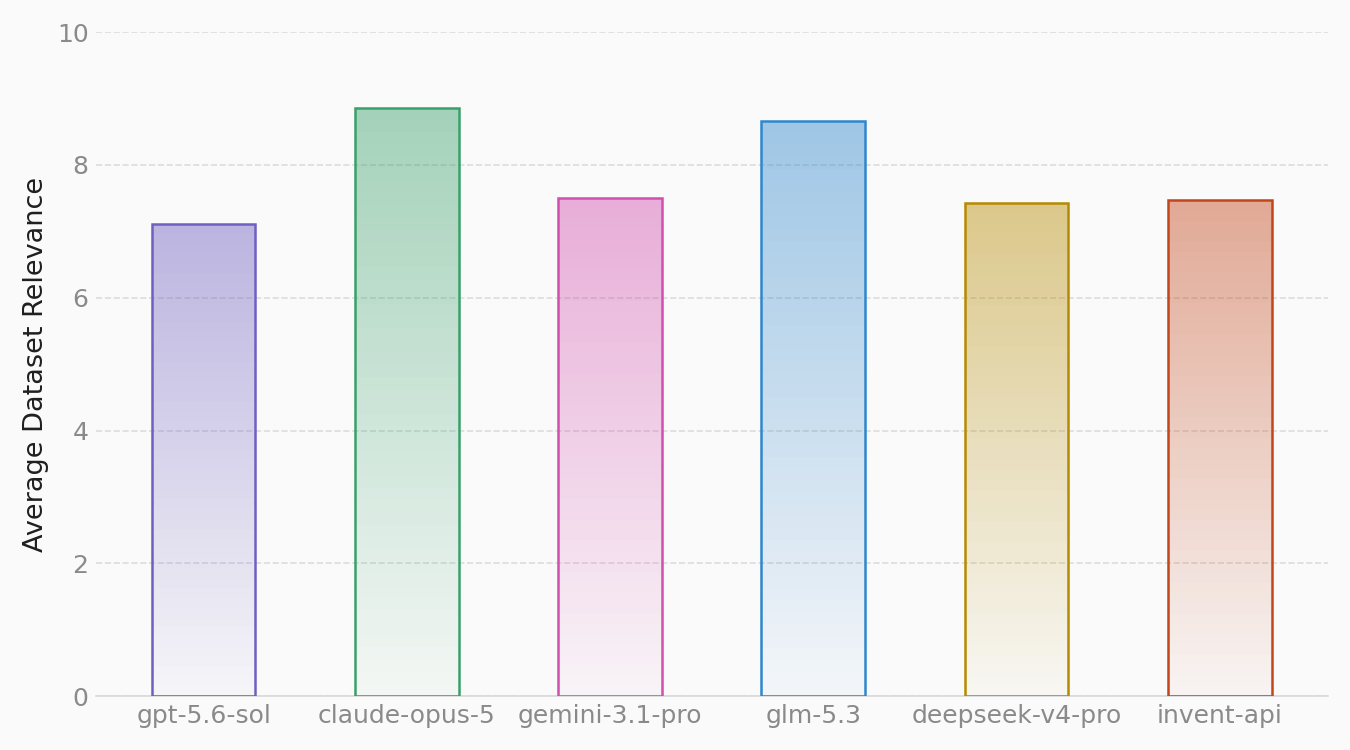}
        \caption{\textbf{Relevance per generator}}
    \label{fig:relevance-per-model}
    \end{subfigure}
    \caption{
       \textbf{Left:} Average dataset prompt and completion quality at the $5{,}000$-sample scale. The \inventapi is the highest-quality dataset generator, ahead of Claude Opus~5 and GLM-5.3. \textbf{Right:} Relevance per generator. Average sample relevance at the $5{,}000$-sample scale. All pipelines produce on-specification data ($7.1$--$8.9$); Claude Opus~5 and GLM-5.3 lead, with the \inventapi competitive.
    }
    \label{fig:post-training}
\end{figure*}

Crucially, the \inventapi's diversity advantage does \emph{not} come at a quality cost, it is pareto-frontier, improving on both objectives at once. Averaged over all datasets at the $5{,}000$-sample scale (Figure~\ref{fig:quality-per-model}), it is the highest-quality
generator at $7.90$, ahead of Claude Opus~5 ($6.76$) and GLM-5.3 ($6.73$) by $\sim\!17\%$, and well above Gemini-3.1-pro ($5.26$) and GPT-5.6 Sol ($5.13$). 

\begin{figure}[h]
    \centering
    \begin{subfigure}[t]{0.48\linewidth}
        \centering
        \includegraphics[width=\linewidth]{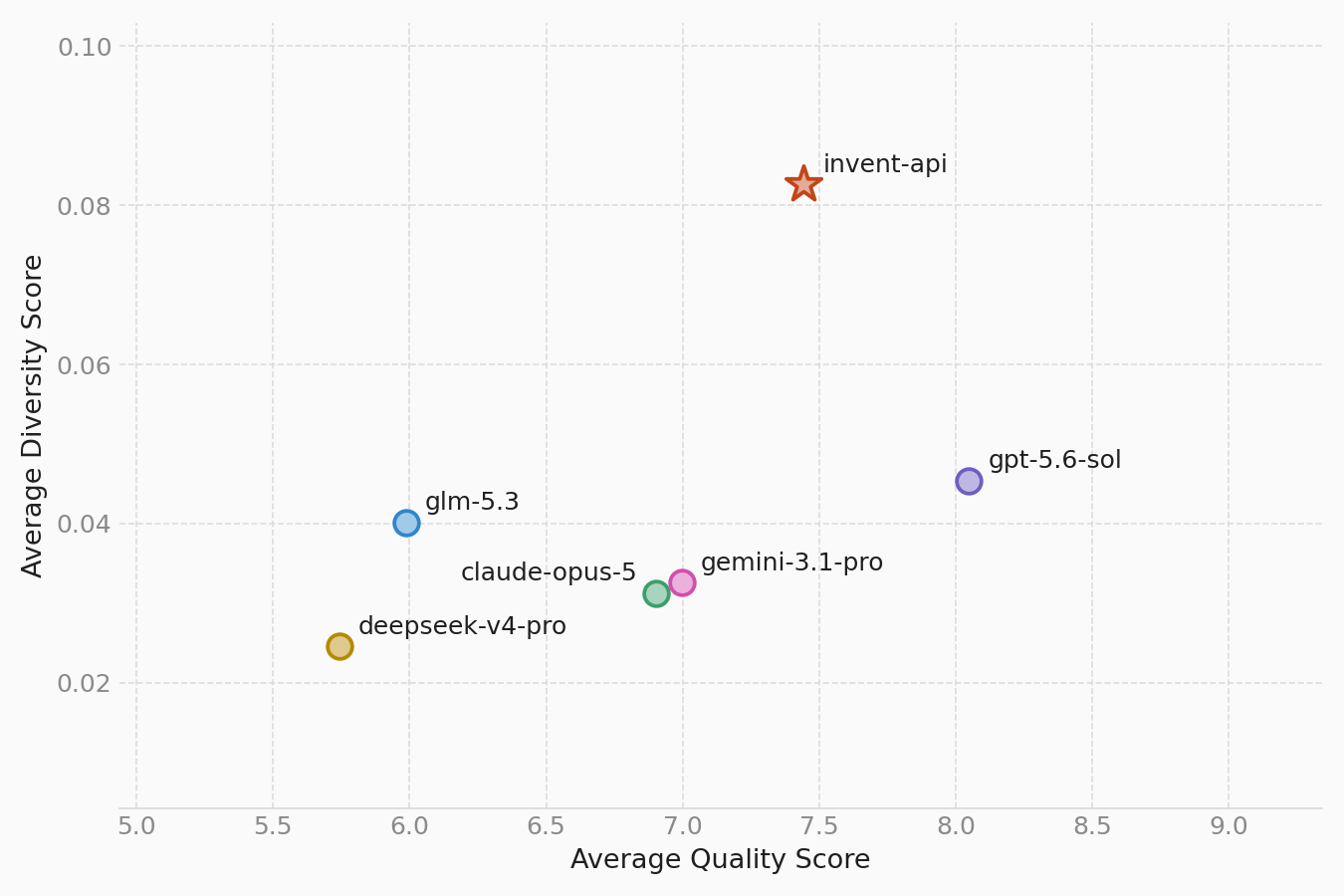}
        \caption{Serbian and Hindi}
        \label{fig:multilingual_sr_hi}
    \end{subfigure}
    \hfill
    \begin{subfigure}[t]{0.48\linewidth}
        \centering
        \includegraphics[width=\linewidth]{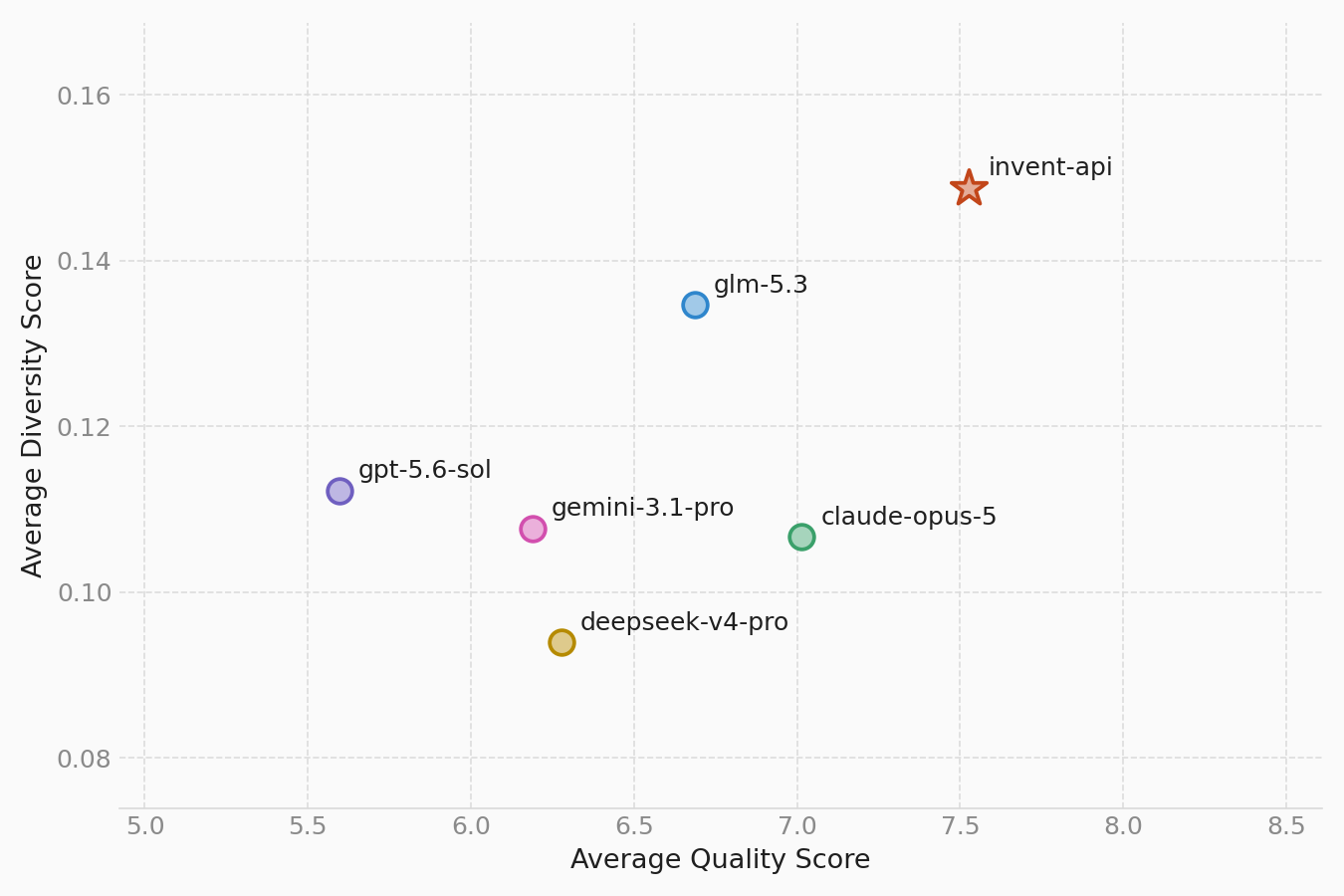}
        \caption{English only}
        \label{fig:multilingual_other6}
    \end{subfigure}
    \caption{\textbf{Diversity vs. Quality in Multilingual Setting.} Diversity--quality trade-off at $5{,}000$ samples for non-English datasets in the constrained setting. In both settings the \inventapi occupies the upper-right region, leading on diversity throughout and on quality for the six additional languages.}
    \label{fig:multilingual}
\end{figure}

\textbf{Diversity and Quality across different languages} \invent supports 244 different languages~\citep{hooker2026expand}, hence it of interest to understand performance across different languages. In Figure~\ref{fig:multilingual} we compare Serbian and Hindi (Figure~\ref{fig:multilingual_sr_hi}), and show the \inventapi is clearly the most diverse ($\approx 0.082$, roughly double the next-best baseline) while retaining high-quality generations; GPT-5.6 Sol reaches slightly higher quality ($\approx 8.05$) but at less than half the diversity, leaving the \inventapi alone in the upper-right. Across six English datasets (Figure~\ref{fig:multilingual_other6}), the separation is cleaner. The \inventapi leads on \emph{both} axes, with the highest quality ($\approx 7.5$) and the highest diversity ($\approx 0.149$, ahead of GLM-5.3 at $0.135$). The multilingual setting therefore reinforces the English results, the \inventapi's diversity advantage is not language-specific. Comparing the \inventapi to itself, quality is nearly unchanged across languages ($\approx 7.5$ in English vs.\ $\approx 7.4$ in Serbian and Hindi), while absolute diversity is lower ($\approx 0.149$ vs.\ $\approx 0.082$); however, every baseline loses more diversity (roughly 60--75\% vs.\ $\approx 44\%$), so the \inventapi's relative advantage widens outside English.

We also measure whether the outputs are generated in the requested language; the results for language confusion \citep{marchisio-etal-2024-understanding} in Hindi are reported in the Appendix~\ref{langid-consistency}. To verify the language of the generated datasets, we applied \texttt{langid}~\citep{lui2012langid} to each generated example. All evaluated baselines exhibit minimal language drift, producing the large majority of examples in the target language; the highest drift rate is 7.25\%, observed for DeepSeek-V4-Pro. \invent{} yields the lowest drift of all methods, with only 0.15\% (less than 1\%) of its Hindi outputs identified as a different language.

\begin{figure}[t]
    \centering
    \includegraphics[width=\linewidth]{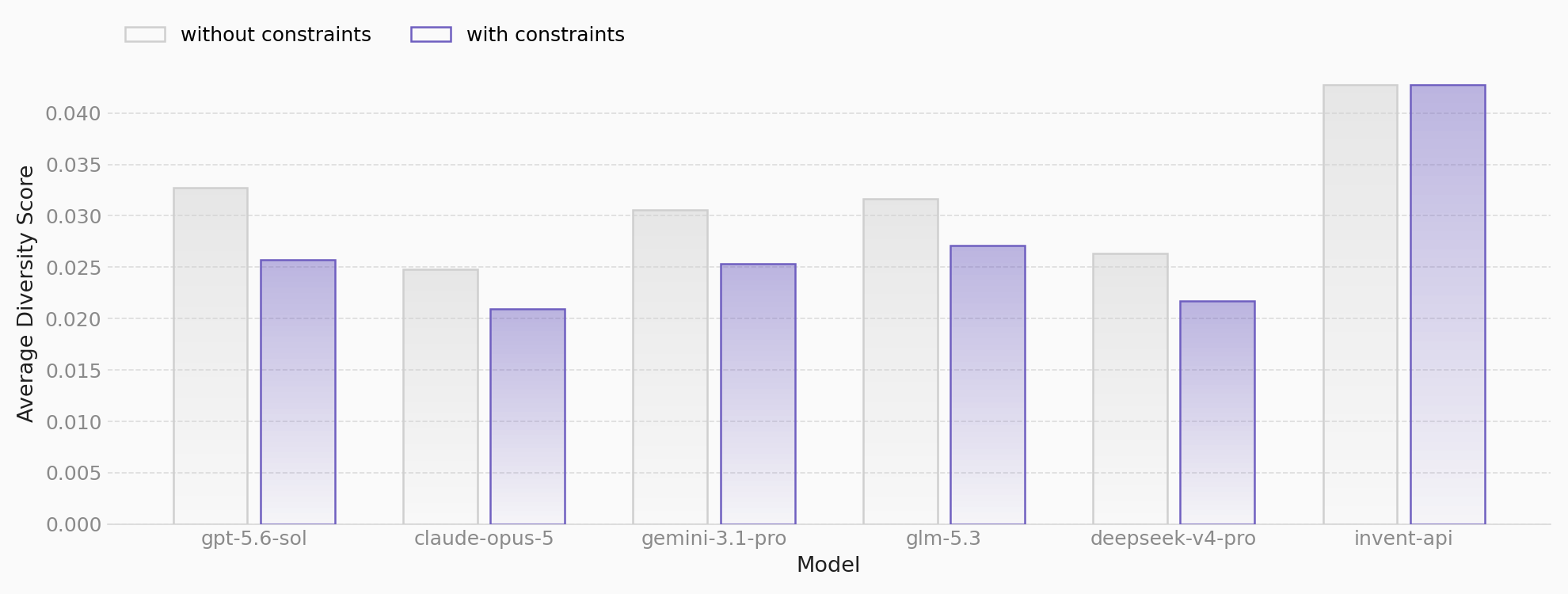}
    \caption{\textbf{Diversity under constrained vs.\ unconstrained settings per generator}, measured with DCScore ($\tau = 0.1$) and and averaged across 8 datasets of 20000 samples each. The \inventapi is the
most robust: its diversity is essentially unchanged ($0.0428$ in both settings), while every baseline drops by $15$--$22\%$. As a result, \emph{under constraints} the \inventapi
has the highest diversity of any pipeline ($0.0428$ vs.\ $0.0271$ for the next best, GLM-5.3).}
    \label{fig:cvu_per_model}
\end{figure}

\begin{figure*}[t]
\centering
    \begin{subfigure}[t]{0.45\textwidth} 
    \centering
        \includegraphics[width=\textwidth]{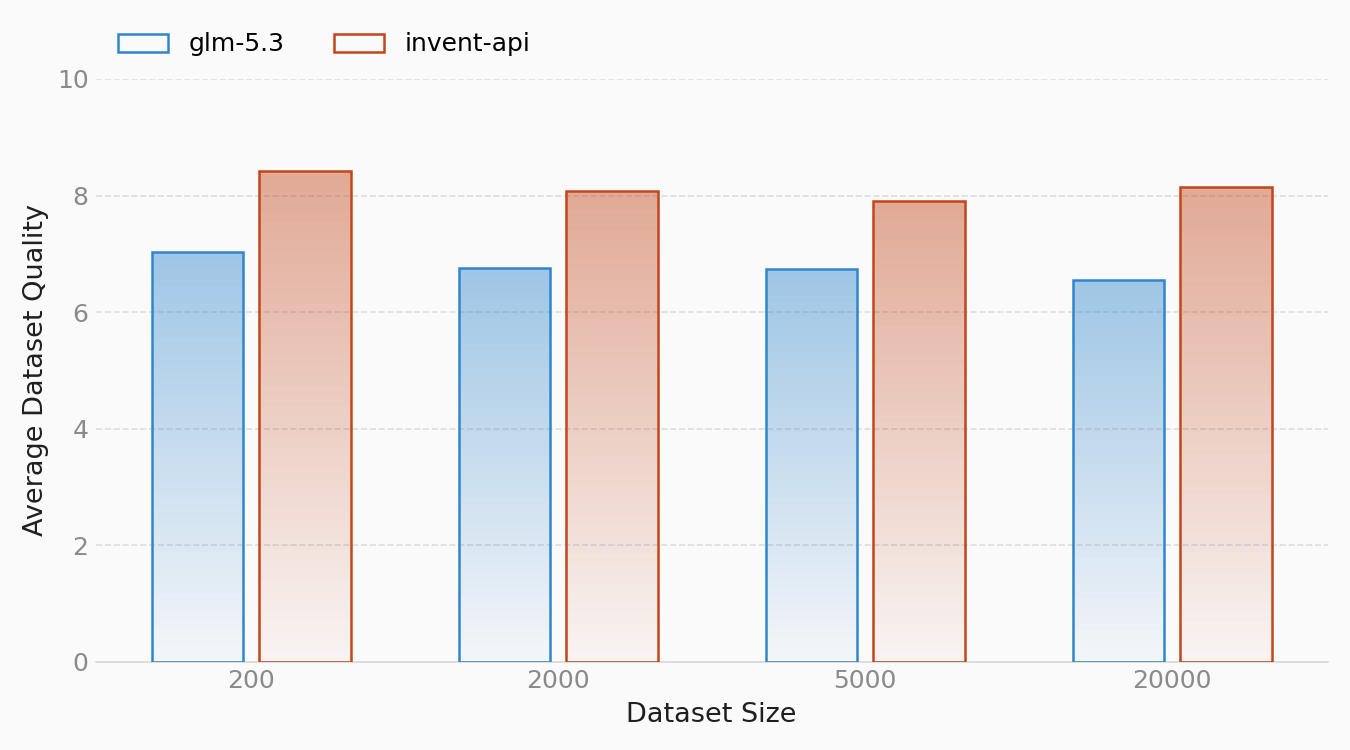}
        \caption{\textbf{Quality} with scale}
    \label{fig:quality-with-scale}
    \end{subfigure}
    \begin{subfigure}[t]{0.45\textwidth} 
    \centering
        \includegraphics[width=\textwidth]{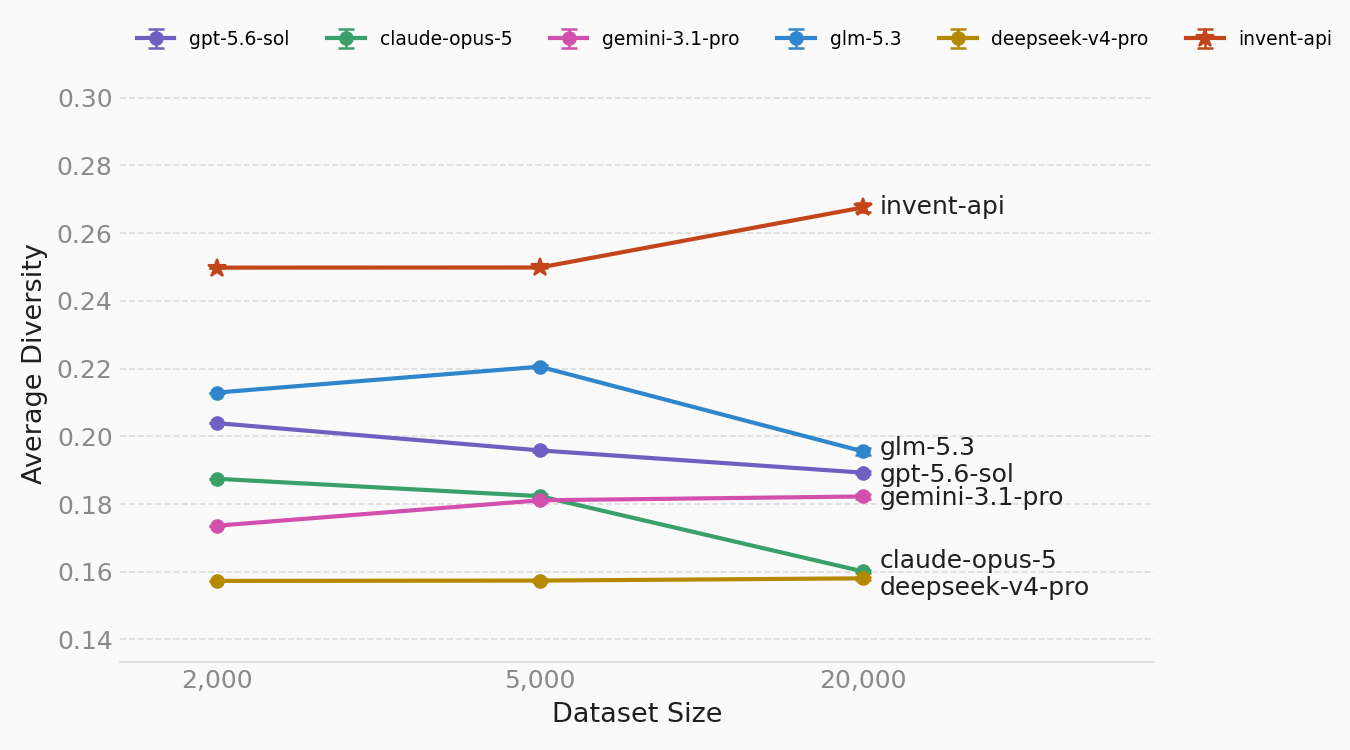}
        \caption{\textbf{Diversity} with scale.}
    \label{fig:diversity-with-scale}
    \end{subfigure}
    \caption{
\textbf{Left:} Quality with scale. Average dataset quality for GLM-5.3 and the \inventapi at $200$, $2{,}000$, and $5{,}000$ and $20{,}000$ samples. The \inventapi leads at every scale and remains stable as the dataset grows. \textbf{Right:} Diversity with scale. Average diversity (DCScore, $\tau=0.1$) for all generators at $2{,}000$, and $5{,}000$ and $20{,}000$ samples measured on $2{,}000$ samples randomly drawn from each dataset. Per-sample diversity declines with scale for most baseline generators, whereas \inventapi{}'s diversity remains high (and even increases), so its relative advantage grows with scale.}
\end{figure*} 

\subsection{Impact of Constraints on Dataset Diversity}

Real-world data requests often impose explicit constraints (length, format, structure, persona), which narrow the space of valid outputs and typically reduce diversity. These constraints typically demand more complex generation patterns which increase difficulty of the synthetic data generation task. Figure~\ref{fig:cvu_per_model} compares
diversity of \textbf{unconstrained} dataset queries versus \textbf{constrained} (Table~\ref{tab:queries_3} and Table~\ref{tab:queries}). Adding constraints reduces diversity for every external baseline, with drops of up to $22\%$. GPT-5.6 Sol drops $22\%$ ($0.0328\!\to\!0.0257$), DeepSeek-v4-Pro $18\%$
($0.0264\!\to\!0.0217$), Gemini-3.1-pro $17\%$ ($0.0306\!\to\!0.0254$), Claude Opus~5 $15\%$ ($0.0248\!\to\!0.0210$) and GLM-5.3 $15\%$ ($0.0317\!\to\!0.0271$). The \inventapi is the
only pipeline whose diversity is unaffected ($0.0428$ in both settings), and no baseline comes close to this robustness.
As a result, \emph{under constraints} the \inventapi has the highest diversity of any pipeline ($0.0428$), about $58\%$ above the next best (GLM-5.3, $0.0271$). Generating data that fulfills precise constraints is critical for real world modelling and it shows that the \inventapi
maintains diversity precisely where constraints would otherwise reduce it.

\subsection{Diversity and Quality with Scale} 
\label{sec:relationship_diversity_and_scale}

\textbf{Diversity degrades with scale.} We also measure how diversity evolves with dataset size. In Figure~\ref{fig:diversity-with-scale} two effects are noticeable. First, absolute per-sample diversity declines for \emph{all} external generators as the dataset grows: this is expected, since each new sample must differ from an ever-larger pool already generated, so a single model prompted the same way increasingly falls back on repeated patterns and the marginal diversity gain per sample steadily shrinks. Second, and more importantly, the \emph{relative} advantage of the \inventapi \textbf{widens} with scale. At 2000 samples, \inventapi already leads the strongest baseline (\texttt{glm-5.3}) by $0.037$ in average diversity ($0.250$ vs.$0.213$, a $17\%$ relative gain), but by 20000 samples this margin grows to $0.072$ ($0.268$ vs. $0.196$), a \textbf{37\%} relative gain that more than doubles the advantage. This widening comes from diverging trends: \inventapi's diversity increases with dataset size, whereas every baseline either plateaus (\texttt{deepseek-v4-pro}, \texttt{gemini-3.1-pro}) or declines (\texttt{glm-5.3}, \texttt{gpt-5.6-sol}, \texttt{claude-opus-5}), suggesting that frontier LLMs increasingly repeat themselves as more data is generated.

\textbf{Quality with scale.} \inventapi's quality lead is stable across scale (Figure~\ref{fig:quality-with-scale}). Against GLM-5.3, \inventapi leads at every size ($8.43$ vs.\ $7.03$ at $200$, $8.09$ vs.\ $6.76$ at $2{,}000$, $7.90$ vs.\ $6.73$ at $5{,}000$, and $8.15$ vs.\ $6.55$ at $20{,}000$), a relative advantage of roughly $17$--$24\%$ throughout. \inventapi's own score stays within a narrow band ($7.90$--$8.43$) as generation scales, while GLM-5.3 declines steadily ($7.03\!\to\!6.55$), indicating that \inventapi's quality holds up even as the dataset grows and diversity pressure increases. Taken together with the diversity results, this shows both diversity and quality persist at scale.

\textbf{Lexical diversity with scale.} As mentioned in section~\ref{sec:eval-metrics}, we complement DCScore based on semantic content with three additional diversity metrics based on textual form. Table~\ref{tab:lexical-prompt} reports them in the \textit{unconstrained} setting at the $2{,}000$, $5{,}000$, and $20{,}000$ dataset sizes, each computed on a fixed $2{,}000$ prompt random sample so the numbers are directly comparable across sizes. The \inventapi is the best generator on \emph{every} metric at \emph{every} scale. Its $n$-gram diversity is the highest and \emph{increases} with dataset size ($1.75\!\to\!1.82\!\to\!1.87$). It's relative gain compared to next closest baseline (GLM-5.3) are 9.4\%, 7.1\%, and 10.7\% at 2K, 5K, and 20K respectively. It also yields the lowest compression ratio at every size ($\approx 3.5$, i.e.\ it has the least repetitive prompts, versus $3.95$--$5.5$ for the baselines). It also has a zero exact-duplicate rate ($0.0\%$) at all scales, while every external baseline outputs $6.2$--$19.2\%$ \emph{exact-duplicate} prompts, rising with dataset size and worst for DeepSeek-V4-Pro. These metrics highlight that \inventapi's prompts are the most varied, the least compressible, and not repetitive, while frontier LLMs increasingly fall back on repetition as more data is generated. Results for the \textit{constrained} setting are reported in Appendix~\ref{app:lexical_constrained}.

\begin{table}[t]
\centering
\small
\setlength{\tabcolsep}{5pt}
\begin{tabular}{lccccccccc}
\toprule
& \multicolumn{3}{c}{NGD$_{\text{prompt}}\uparrow$} & \multicolumn{3}{c}{CR$_{\text{prompt}}\downarrow$} & \multicolumn{3}{c}{Dup$_{\text{prompt}}$ (\%)$\downarrow$} \\
\cmidrule(lr){2-4}\cmidrule(lr){5-7}\cmidrule(lr){8-10}
Data Variant & 2K & 5K & 20K & 2K & 5K & 20K & 2K & 5K & 20K \\
\midrule
\inventapi & \textbf{1.75} & \textbf{1.82} & \textbf{1.87} & \textbf{3.67} & \textbf{3.51} & \textbf{3.47} & \textbf{0.0} & \textbf{0.0} & \textbf{0.0} \\
GLM-5.3 & 1.57 & 1.70 & 1.69 & 4.70 & 3.95 & 3.96 & 6.2 & 6.8 & 8.9 \\
Gemini-3.1-pro & 1.53 & 1.62 & 1.66 & 4.59 & 4.22 & 4.08 & 7.5 & 6.8 & 6.5 \\
GPT-5.6 Sol & 1.60 & 1.65 & 1.59 & 4.33 & 4.33 & 4.39 & 9.6 & 10.4 & 12.4 \\
Claude Opus~5 & 1.55 & 1.54 & 1.48 & 4.27 & 4.25 & 4.28 & 8.5 & 7.9 & 8.4 \\
DeepSeek-V4-Pro & 1.07 & 1.07 & 1.07 & 5.50 & 5.53 & 5.51 & 19.2 & 18.5 & 18.8 \\
\bottomrule
\end{tabular}
\caption{Lexical prompt diversity and duplication in the \emph{unconstrained} setting, averaged over
the eight datasets. Each dataset size (2K/5K/20K) is measured on a fixed $2{,}000$ prompt random sample,
so columns are directly comparable. n-gram diversity (NGD): 
$\uparrow$ implies more diverse. Compression Ration (CR):
$\downarrow$ implies less redundant; Dup is the percentage of exact-duplicate prompts ($\downarrow$ is better). Best data variant is indicated in bold for each column.}
\label{tab:lexical-prompt}
\end{table}

\subsection{Relevance of Generated Samples}

\textbf{Relevance of Generated Samples} Relevance measures whether a sample matches the dataset specification, independent of how well-constructed or diverse it is. Figure~\ref{fig:relevance-per-model} reports average relevance per model at $5{,}000$ samples. While relevance is high across all generators ($7.1$--$8.9$), confirming that every generator produces on-specification data, the \inventapi sits in the middle rather than at the top. Claude Opus~5 ($8.86$) and GLM-5.3 ($8.66$) outperform \inventapi by more than a point. The \inventapi ($7.47$) is comparable to Gemini-3.1-pro ($7.51$),  DeepSeek-v4-pro ($7.44$) and above GPT-5.6 Sol ($7.11$). \inventapi's current advantage lies in maintaining diversity and quality but currently it does not always produce the most on-specification relevant samples. Improving relevance without sacrificing diversity remains an important future improvement area for the \inventapi. 

\subsection{Impact on downstream post-training quality}

To test whether better generated data produces a better model,  we repeat the following protocol for three models: Llama-3.3-70B, Gemma-4-31B-it and Qwen3.5-9B. For each, we compare the un-tuned base model against five independently fine-tuned variants, each trained on a 20K \textbf{Medical QA} dataset produced using a different generator. All six models answer the same set of held-out medical prompts. We use an LLM judge to rank all six responses in a single call from best to worst. 

\begin{figure}
    \centering
    \begin{subfigure}[t]{0.48\linewidth}
        \centering
        \includegraphics[width=\linewidth]{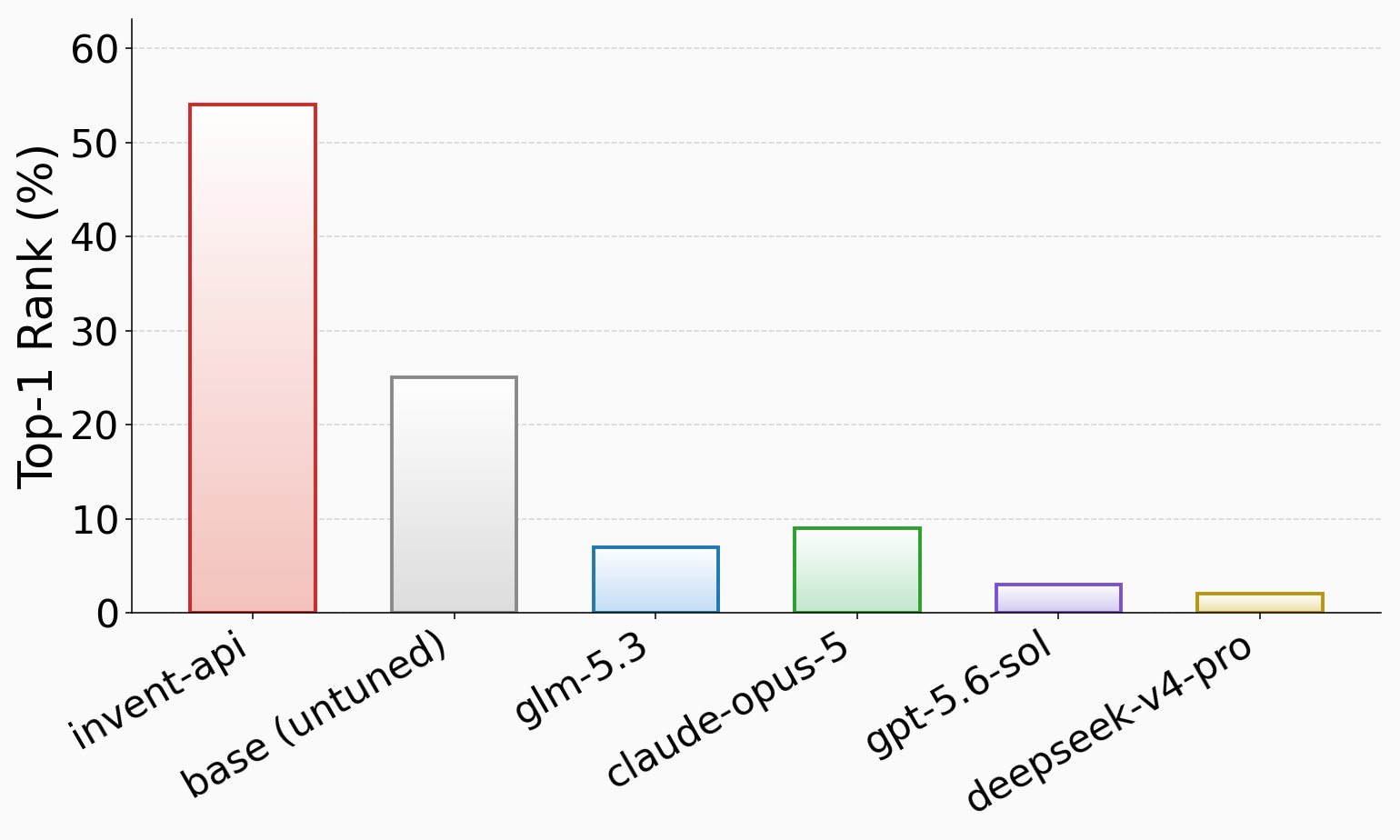}
        \caption{Llama-3.3-70B}
        \label{fig:ranking-llama}
    \end{subfigure}\hfill
    \begin{subfigure}[t]{0.48\linewidth}
        \centering
        \includegraphics[width=\linewidth]{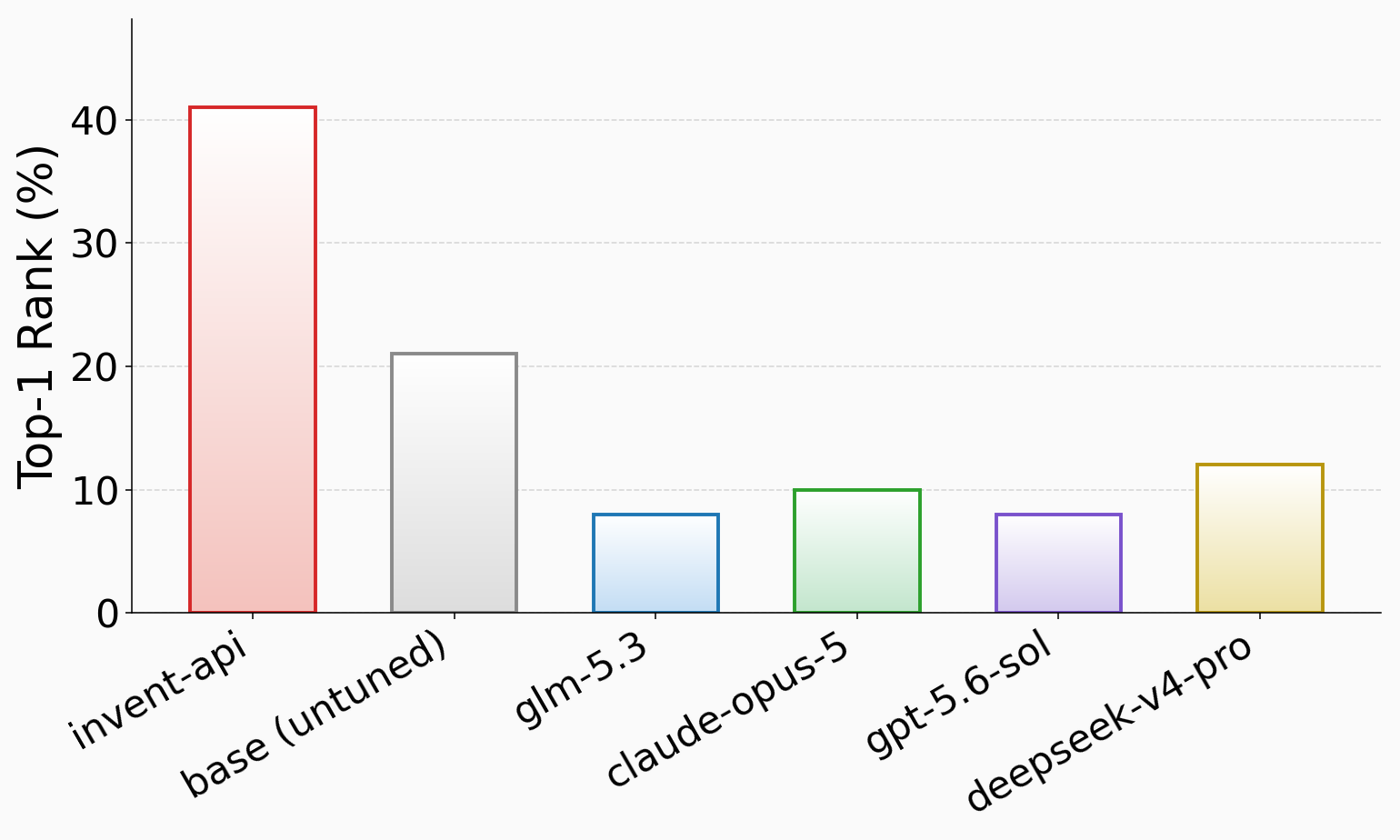}
        \caption{Gemma-4-31B-it}
        \label{fig:ranking-gemma}
    \end{subfigure}
    \caption{\textbf{Top-1 rank share in downstream ranking evaluation.} Percentage of held-out medical test prompts on which each fine-tune is ranked first, with all six responses to a prompt judged together in a single call. Within each panel, the x-axis lists models fine-tuned on different synthetic data variants, plus the untuned base model (grey) as a control. \textbf{Left:} Llama-3.3-70B. The \inventapi fine-tune is the clear winner, ranking first on $54\%$ of prompts, more than double the untuned base at $25\%$. \textbf{Right:} Gemma-4-31B-it. The \inventapi fine-tune again leads at $41\%$, well above the base at $21\%$. In both cases, the models fine-tuned on data from other generators (GLM-5.3, Claude Opus~5, GPT-5.6 Sol, DeepSeek-V4-Pro) trail the base by a wide margin.}
    \label{fig:ranking}
\end{figure}

Across all models we finetune, we observe that models trained on \inventapi data outperform by a large margin. As seen in Figure~\ref{fig:ranking} (left) Llama-3.3-70B the model trained on \inventapi data is ranked first on 54\% of the samples. This is more than double the base un-tuned model (25\%) and six times the strongest competing fine-tuned model trained on Claude Opus 5 generated data (9\%). These patterns hold for Gemma-4-31B-it (Figure~\ref{fig:ranking}, right), the four models fine-tuned on data from other generators all fail to beat the base for Top-1 rank share (DeepSeek-V4-Pro (12\%), Claude Opus 5 (10\%), GLM-5.3(8\%) and GPT-5.6 Sol (8\%)). For Qwen3.5-9B, the untuned base is a really strong starting point and edges the \inventapi fine-tune on Top-1 rank share ($44\%$ vs.\ $38\%$), yet \inventapi still ranks above every other generator's fine-tune (see Appendix~\ref{app:qwen-ranking} for details).
These results collectively confirm that differences in data quality and diversity carry through to the models trained on that data.  The ordering is consistent with the dataset-level results in Sections~\ref{sec:overall_performance} and~\ref{sec:relationship_diversity_and_scale}. 

\section{Related Work}

Historically, significant effort has been invested in collecting, labeling, filtering, and reshaping data to fit a training objective \citep{wang2024survey} A large body of work has emerged which shows that efforts to better curate training corpus, including de-duping \citep{taylor2022galactica, kocetkov2211stack}, data pruning \citep{marion2309less,singh2024aya,sorscher2022beyond,albalak2024survey,tirumala2023d4,chimoto2024critical} or data prioritization \citep{boubdir2023prompts,thakkar2023self} can lead to better model quality. However, most of these works start from a existing data pool and aim to modify it to arrive at higher quality data. The result is often a dataset that only approximates the desired behavior, and a model trained around the limitations of the available data rather than the target capability. In contrast, our work is focused explicitly on the zero data regime where no data is available and optimizing directly for any set of constraints or capabilities a practitioner desires.

\textbf{Dataset search approaches} Dataset search poses unique challenges distinct from traditional web search \citep{brickley2019dataset}. Users span a wide range of expertise and goals, and in many cases those goals are far removed from the datasets themselves. The datasets are also often difficult to peruse manually. Despite advances in interpreting natural-language intents \citep{chapman2020datasearch}, users still struggle with incomplete and inconsistent metadata, with expressing information-seeking needs as structured search constraints, and with assessing dataset relevance \citep{kacprzak2019query, li2025surveyopendatasetsearch}. Crucially, dataset search is fundamentally limited to \emph{retrieving} data that already exists; when no suitable dataset is available, search offers no recourse.

\textbf{Synthetic data} Synthetic data offers several compelling advantages. It allows practitioners to target long-tail scenarios that are rare in the real distribution \citep{wang2024survey, long2024llmsdriven}, minimizes the cost of additional data collection, and can self-recursively increase the difficulty of examples to elicit more advanced capabilities \citep{xu2024wizardlm, luo2023wizardcoder}. The cost of generating new data has decreased over time, allowing for “on-the-fly” optimization in the data space. \citep{shimabucoro-etal-2024-llm}. These benefits, however, are contingent on the diversity of the generated data. Training on narrow or repetitive synthetic samples degrades performance and, in the recursive limit, induces \emph{model collapse}, in which successive models progressively lose the tails of the distribution \citep{shumailov2024collapse, schaffelder-gatt-2026-synthetic}. Maintaining diversity is therefore a central objective for any synthetic data generation method. Most data synthesis approaches leverage existing data structures while adapting them to new requirements. Seed-based methods such as Self-Instruct \citep{wang2023selfinstruct} and Alpaca \citep{taori2023alpaca} bootstrap large instruction-tuning sets from a small pool of human-written seed examples, while evolutionary methods such as Evol-Instruct/WizardLM \citep{xu2024wizardlm} and its code variant WizardCoder \citep{luo2023wizardcoder} iteratively rewrite an existing dataset into more complex and diverse instances. Other work tailors synthetic data to a target model or task, but still conditions on seed instructions or an existing corpus \citep{wang2024codeclm}.  In contrast, our work assumes the end user has \emph{no} access to any existing data. We construct a dataset zero-shot, from a natural-language description of the desired task alone. 

\section{Conclusion}
\label{sec:conclusion}

A systematic weakness of using external model APIs directly is that their generated samples fall into repetitive patterns, which erodes the diversity of the resulting dataset. This effect is clearest when tracking diversity across dataset sizes. When a single model is prompted repeatedly, novel outputs become increasingly hard to produce and per-sample diversity steadily declines. \inventapi is far more resistant to this collapse, and its advantage is evident as we scale across dataset size. At scale, it generates data that is substantially more diverse than every baseline while maintaining the highest overall dataset quality and remaining on-specification. 

\section*{Limitations}

There are some important shortcomings of \invent. Today it produces instruction and preference datasets for SFT and alignment training but does not yet extend to tool-call traces or multi-step, agentic trajectories a model produces when driving a harness. We leave this as the subject of future work as harness datasets tend to be more particular to a given environment: the available tools, their signatures and the harness's control flow. That coupling makes such datasets narrower and faster to go stale as an environment evolves, and harder to reuse across settings than regular SFT and preference datasets supported at present. There is also a valid question of whether this data is best internalized in parametric weight updates. Additionally, \invent is limited to text modality at present. In the future, we hope to extend this to support multimodal dataset generation. 

\bibliography{custom, anthology}

\appendix

\section{Appendix}

\subsection{Quality Score Description}
\label{app:quality-score-description}
\textbf{Prompt quality.} This rubric grades the generated prompt on a $0$--$10$  scale based on criteria such as specificity, scope, and completeness. Low scores ($1$--$4$) denote incoherent, vague, or under-specified requests; mid scores ($5$--$6$) a clear, executable task lacking depth or context; and high scores ($7$--$10$) well-structured prompts with explicit success criteria. Crucially, prompts are judged \emph{within} the form the dataset request specifies, so a deliberately short prompt is not penalized for brevity it was asked to have by the user.

\textbf{Completion Quality}: grades  the generated response based on a $0$--$10$ based on its overall quality, structure, and adherence to the prompt's stated constraints. Low scores ($1$--$2$) denote incoherent, incorrect, or incomplete answers; mid scores ($3$--$5$) a competent, accurate response covering the main points but lacking depth; and high scores ($6$--$10$) comprehensive, well-organized answers that address caveats and edge cases. As with prompt quality, depth is judged relative to any length, format, or style limits the dataset request imposes.

Below we show full judge prompt templates:

\label{app:quality-judge-template}
\newtcolorbox{titlecard}[1]{
  enhanced, breakable,
  colback=cardbg, colframe=cardframe, boxrule=0.5pt, arc=4pt,
  coltitle=white, fonttitle=\bfseries,
  colbacktitle=titleblue,
  attach boxed title to top left={xshift=10pt, yshift=-\tcboxedtitleheight/2},
  boxed title style={colframe=titleblue, colback=titleblue, arc=3pt,
                     boxrule=0pt, top=2pt, bottom=2pt, left=6pt, right=6pt},
  title={#1},
  top=12pt, left=9pt, right=9pt, bottom=9pt
}

\newtcolorbox{promptbox}[1]{
  enhanced, breakable,
  colback=white, colframe=sectionframe, boxrule=0.5pt, arc=3pt,
  colbacktitle=sectionbg, coltitle=black, fonttitle=\bfseries,
  title={#1},
  left=9pt, right=9pt, top=7pt, bottom=7pt,
  toptitle=3pt, bottomtitle=3pt, lefttitle=9pt,
  before skip=8pt, after skip=8pt
}

\newlist{scale}{itemize}{1}
\setlist[scale]{leftmargin=1.4em, itemsep=3pt, topsep=3pt, parsep=0pt, label={}}

\setlength{\parindent}{0pt}
\setlength{\parskip}{4pt}

\begin{titlecard}{Quality and Relevance Judge Template}
A strict, meticulous judge for curating supervised fine-tuning datasets. Given a
\textbf{target specification} (a dataset-level data request plus language tags) and one
\textbf{candidate} prompt/completion pair, the judge decides whether the candidate belongs
in the dataset and how good it is.
\begin{itemize}[leftmargin=1.4em, itemsep=2pt, topsep=4pt]
  \item \textbf{Scoring axes:} relevance, prompt quality, completion quality --- each an integer $0$--$10$.
  \item \textbf{Output:} a single compact JSON object, no other text.
  \item \textbf{Template placeholders:} \ph{query}, \ph{language}, \ph{prompt}, \ph{completion}, \ph{prompt\_quality\_rubric}, \ph{completion\_quality\_rubric}.
\end{itemize}
\end{titlecard}

\begin{promptbox}{System Message}
You are a strict, meticulous judge who curates supervised fine-tuning datasets. Given a
\textsc{target specification} (a dataset-level data request and language tags) and one
\textsc{candidate} prompt/completion pair, you decide whether the candidate genuinely
\textsc{belongs} in that dataset and how good it is. Be decisive and consistent: reward
concrete, self-contained, on-topic prompts paired with accurate, complete completions;
penalize off-topic, wrong-language, incomplete/truncated, answer-shaped, or meta
(\enquote{generate a dataset of\ldots}) prompts and shallow, wrong, or evasive completions.
You score the candidate on \textsc{three independent axes} --- relevance (does the sample
belong in the dataset?), prompt quality (how good is the prompt on its own?), and completion
quality (how good is the completion as an answer to the prompt?) --- each $0$--$10$. Reply with
\textsc{only} a single compact JSON object and no other text.
\end{promptbox}

\begin{promptbox}{Prompt Quality Rubric}
Assign a quality score from $0$ to $10$ for the overall quality of the prompt.

\smallskip
\textbf{Scale}
\begin{scale}
  \item \textbf{0 (Empty prompt)} --- There is no prompt. It is a null entry.
  \item \textbf{1--2 (Incoherent / Nonsensical)} --- The request is unreadable, random, or missing
        meaning. Single words. Gibberish, unrelated fragments, no clear task. May be impossible to
        answer given the available context. No discernible task, goal, or meaning.
        Examples: \enquote{asdf quantum banana}, \enquote{Help}, \enquote{AI}, \enquote{Write something}, \enquote{???}
  \item \textbf{3--4 (Unclear or Under-Specified)} --- Readable but lacks sufficient detail to execute
        effectively. Identifiable topic but missing critical parameters that specify scope, audience,
        format or constraints. Overly general request.
        Examples: \enquote{Tell me about technology}, \enquote{Write about climate change},
        \enquote{Explain machine learning}, \enquote{What do you think?}
  \item \textbf{5--6 (Basic Clarity)} --- Identifiable task with clear action verb. Straightforward but
        lacks depth or strategic context. Examples: \enquote{Summarize this article},
        \enquote{Write a 3-paragraph explanation for beginners}, \enquote{List 5 pros and cons of remote work}
  \item \textbf{7--8 (Well-Structured)} --- Includes relevant context or use case. Clear success criteria
        or deliverable specifications. May request comparative analysis or structured reasoning.
        Example: \enquote{Write a 150-word summary of this AI safety paper for a tech newsletter audience,
        emphasizing practical implications}
  \item \textbf{9--10 (Expert-Level / Meta-Optimized)} --- Advanced prompt engineering with sophisticated
        scaffolding. Explicit role assignment or persona specification. Multi-dimensional reasoning or
        evaluation criteria. Anticipates edge cases, nuance, or quality standards. May include
        self-correction mechanisms or iterative instructions.
\end{scale}

\smallskip
\textbf{Special Cases}
\begin{itemize}[leftmargin=1.4em, itemsep=4pt, topsep=3pt]
  \item \textbf{Respect dataset-request constraints:} Do \textsc{not} penalize a prompt for lacking detail,
        length, context, or scaffolding that the dataset request explicitly told it to omit or cap. When the
        request mandates a form --- e.g., short or one-line prompts, no provided context, a fixed template, or
        a plain/casual register --- score how well the prompt is crafted \emph{within} that form: is it clear,
        purposeful, and complete for its intended scope? A deliberately terse prompt that cleanly specifies its
        task can still score $7+$; reserve $3$--$4$ for prompts that are vague or under-specified \emph{beyond}
        what the request requires. Judge specification quality relative to the requested form, not against an
        unconstrained ideal.
  \item \textbf{Multi-turn conversation:} if the prompt is a chat involving a multi-turn conversation, score
        based on the last turn in the conversation.
  \item \textbf{A complex prompt that is poorly asked:} Focus on prompt structure and specification, not topic
        complexity. A difficult question should still be ranked low for quality if it is poorly asked or if the
        desired output is not clear.
  \item \textbf{Long prompts are not necessarily high quality:} Watch out for verbose but unfocused prompts
        masquerading as detailed. If a request is needlessly lengthy, or lacks helpful structure, it should be
        scored as low quality.
\end{itemize}
\end{promptbox}

\begin{promptbox}{Completion Quality Rubric}
Assign a quality score from $0$ to $10$ for the completion \textsc{as an answer to the prompt}.

\smallskip
\textbf{Scale}
\begin{scale}
  \item \textbf{0 (Empty completion)} --- There is no response; it is a null value.
  \item \textbf{1 (Incoherent / Nonsensical)} --- Unreadable response, completely wrong answer, or fails to
        address the prompt. Gibberish, broken text, or semantically empty output. Completely misunderstands or
        ignores the prompt. Addresses an entirely wrong topic or provides irrelevant information. Contains
        critical factual errors that make the response unusable. May acknowledge the prompt but provides no
        actual answer.
  \item \textbf{2 (Inadequate or Incomplete)} --- Addresses the prompt but with major gaps or issues. Provides
        only trivially basic or superficial information. Answers some parts of the prompt while ignoring others.
        Contains significant accuracy issues or misleading statements. Missing critical context, details, or
        qualifications. Oversimplified to the point of being unhelpful. Disorganized or poorly structured.
  \item \textbf{3--5 (Basic Clarity)} --- Competent response that meets basic requirements. Directly addresses
        the prompt. Factually accurate with no major errors. Reasonably complete coverage of main points. Clear
        and understandable structure. May include basic examples or context. Sufficient for basic understanding
        but lacks depth or nuance.
  \item \textbf{6--8 (Well-Structured)} --- Comprehensive coverage with relevant details and context. Deep,
        nuanced understanding with multiple perspectives. Includes helpful examples, analogies, or illustrations.
        Well-organized with clear logical flow. Addresses edge cases, caveats, limitations, or trade-offs.
        Anticipates follow-up questions or potential misconceptions. Provides practical, actionable guidance.
  \item \textbf{9--10 (Expert-Level / Meta-Optimized)} --- Goes meaningfully beyond what was asked (in helpful
        ways). Demonstrates mastery with strategic depth and synthesis. Provides novel insights, non-obvious
        connections, or transferable frameworks. Identifies unstated assumptions or addresses the underlying
        need. Structured for maximum clarity and usability (may include tables, schemas, decision trees).
        Anticipates and prevents common mistakes or misunderstandings. Teaches underlying principles, not just
        answers the prompt.
\end{scale}

\smallskip
\textbf{Special Cases}
\begin{itemize}[leftmargin=1.4em, itemsep=4pt, topsep=3pt]
  \item \textbf{Respect dataset-request constraints:} Do \textsc{not} penalize a completion for lacking depth,
        length, elaboration, examples, or nuance that the dataset request explicitly excluded or capped. When the
        request mandates the completion's length, format, style, or depth (e.g., \enquote{one-sentence answers},
        \enquote{terse}, \enquote{bullet list only}, \enquote{no explanation}), score how fully and accurately it
        satisfies the prompt \emph{within} those bounds. A concise, correct, well-executed answer that obeys a
        brevity or format requirement can still score $6+$ (or higher if it executes the constrained form
        excellently); do not treat requested brevity or simplicity as \enquote{inadequate} or \enquote{superficial}.
        Judge depth relative to the requested form, not against an unconstrained ideal.
  \item \textbf{Multi-turn conversation:} score the completion as a response to the last user turn.
  \item \textbf{Long completions are not necessarily high quality:} verbose, padded, or repetitive answers that
        do not add substance should be scored as low quality.
\end{itemize}
\end{promptbox}

\begin{promptbox}{Complete Judge Prompt}
\textbf{Target specification}
\begin{itemize}[leftmargin=1.4em, itemsep=2pt, topsep=3pt]
  \item Dataset data request (task \& topic; applies to the whole dataset): \ph{query}
  \item Target language: \ph{language}
\end{itemize}
Score the candidate sample containing prompt and completion on \textsc{three independent axes}, each an
integer $0$--$10$.

\tcbline
\textbf{Axis 1 --- \code{relevance\_score}}

Does the candidate \textsc{sample} belong in the target dataset? Membership requires \textsc{four hard
criteria}, all of which must hold. They are listed in order of importance --- (A) and (B) are foundational,
and (C) outranks (D): a clear violation of an explicitly stated constraint is a more serious membership
failure than a prompt-usability defect.
\begin{itemize}[leftmargin=1.6em, itemsep=3pt, topsep=3pt]
  \item \textbf{(A) Task \& Topic} --- the sample matches the dataset request's task type \emph{and} subject matter.
  \item \textbf{(B) Language} --- the sample is written in the target language the request specifies.
  \item \textbf{(C) Constraint Following} --- the sample obeys the constraints the dataset request
        \textsc{explicitly} states for the prompt and/or completion (e.g.\ length, format, structure, style,
        persona). Only constraints the request actually states count; if the request states no constraints,
        treat (C) as satisfied --- do \textsc{not} invent constraints.
  \item \textbf{(D) Usable Prompt} --- the prompt is a self-contained, complete, standalone user prompt:
        \textsc{not} an answer/completion, \textsc{not} a truncated fragment, \textsc{not} a meta-instruction
        about building or generating a dataset, \textsc{not} off-topic filler, and \textsc{not} dependent on
        hallucinated or absent context (e.g.\ referring to a file, document, image, table, link, or earlier
        conversation that is not actually included in the prompt).
\end{itemize}

\textbf{\code{relevance\_score} scale}
\begin{scale}
  \item \textbf{0 (Empty)} --- No prompt or completion / null / unreadable.
  \item \textbf{1--2 (Does not belong)} --- Fails (A) or (B): off-topic for the dataset request's task/subject,
        wrong task type, or wrong language. E.g.\ target $=$ customer-support inquiries, candidate $=$
        \enquote{The mitochondria is the powerhouse of the cell.}
  \item \textbf{3 (Constraint violation)} --- On-topic and in-language, but clearly violates an explicitly
        stated (C) constraint (e.g.\ wrong required format, outside a stated length, missing a mandated
        structure). Because (C) outranks (D), score this \textsc{lower} than a mere usability defect.
  \item \textbf{4 (Unusable prompt)} --- On-topic, in-language, and not in clear violation of a stated
        constraint, but fails (D): it is an answer/completion rather than a prompt, a bare fragment or gibberish,
        a truncated/incomplete request, a meta / data-generation instruction (e.g.\ \enquote{Generate 50 billing
        dispute tickets}), or it relies on absent/hallucinated context (e.g.\ \enquote{summarize the attached
        file}, \enquote{based on the table above}).
  \item \textbf{5--6 (Borderline member)} --- Meets (A), (B), (D) and commits no outright (C) violation, but
        follows the stated constraints only \textsc{partially} or loosely, and/or is generic, under-specified,
        or only tangential to the requested subject.
  \item \textbf{7--8 (Belongs)} --- Meets (A), (B), (D) and fully follows every stated constraint (C): clearly
        matches task $+$ topic $+$ language, satisfies all requested constraints, and is a complete, concrete,
        standalone prompt with enough specificity to be a solid example. Minor gaps only.
  \item \textbf{9--10 (Excellent fit)} --- Meets all four criteria, follows every stated constraint precisely
        and completely, and is realistic, specific, and fully self-contained --- an ideal member of this dataset.
\end{scale}

\textbf{Relevance notes.} First read the free-form dataset request to extract (A) task $+$ topic, (B) any
language requirement, and (C) any \textsc{explicit} form/length/format/style/persona constraints. Priority
order is (A)/(B) $>$ (C) $>$ (D): wrong language or topic caps at $1$--$2$; a clear violation of a stated
constraint (C) caps at $\leq 3$; a usability defect (D) caps at $\leq 4$. Full compliance with all stated
constraints is required to score $7+$, and partial/loose compliance caps at $5$--$6$. If the request states
no constraints, (C) is satisfied --- do not cap the score for it and do not invent constraints. For
multi-turn candidates, judge the whole conversation, anchored on the user's request(s).

\tcbline
\textbf{Axis 2 --- \code{prompt\_quality\_score}}

How good is the candidate \textsc{prompt} on its own, independent of the dataset it targets? Use the
\textbf{Prompt Quality Rubric} above (\ph{prompt\_quality\_rubric}; score the user prompt only).

Prompt quality is \textsc{independent} of relevance: an off-topic prompt can still be well-written (high
quality, low relevance), and a perfectly on-topic prompt can be sloppy or under-specified (high relevance,
low quality). Score the two axes separately. However, do \textsc{not} penalize prompt quality for a property
the dataset request explicitly requires. If the request mandates a form --- e.g., short/one-line prompts, no
supplied context, a fixed template, plain casual phrasing, a specific length or format --- judge how well the
prompt executes within that form, not whether it has the depth, length, or scaffolding that would score higher
for an unconstrained prompt. A prompt that is concise because the request demanded concision is not
\enquote{under-specified.}

\tcbline
\textbf{Axis 3 --- \code{completion\_quality\_score}}

How good is the candidate \textsc{completion} as an answer to the candidate prompt? Use the
\textbf{Completion Quality Rubric} above (\ph{completion\_quality\_rubric}; score the completion only, in
light of what the prompt asks for and the dataset request's expectations).

Completion quality is \textsc{independent} of the other two axes: a great answer to an off-topic or sloppy
prompt still scores high here, and a weak answer to an excellent prompt scores low. However, do \textsc{not}
penalize completion quality for a property the dataset request explicitly requires. If the request mandates
the completion's length, format, style, or depth (e.g., \enquote{one-sentence answers,} \enquote{terse,}
\enquote{bullet list only,} \enquote{no explanation}), judge how well the completion fulfills the prompt within
those bounds, not whether it has the elaboration, examples, or nuance that would score higher for an
unconstrained answer.

\tcbline
\textbf{Decision}

Set \code{relevant} to \enquote{yes} \textsc{only} if (A) \textsc{and} (B) \textsc{and} (C) \textsc{and} (D)
all hold --- this always corresponds to \code{relevance\_score} $\geq 5$; otherwise set it to \enquote{no}
(\code{relevance\_score} $\leq 4$). (C) \enquote{holds} when the sample commits no clear violation of any
explicitly stated constraint; (D) \enquote{holds} when the prompt is usable as defined above. Keep
\code{relevance\_score} and the yes/no decision consistent.

\tcbline
\textbf{Return \textsc{only} this JSON object and nothing else}

\begin{tcolorbox}[colback=cardbg, colframe=cardframe, boxrule=0.5pt, arc=2pt,
                  left=6pt, right=6pt, top=5pt, bottom=5pt, fontupper=\ttfamily\small]
\{"relevance\_score": \textless int 0--10\textgreater, "prompt\_quality\_score": \textless int 0--10\textgreater,
"completion\_quality\_score": \textless int 0--10\textgreater, "relevant": \textless "yes" or "no"\textgreater,
"reason": \textless string $\leq$ 30 words naming the deciding criteria for relevance, prompt quality and
completion quality\textgreater\}

\end{tcolorbox}

---------------------------------------------------------------

{\ttfamily\small CANDIDATE PROMPT:} \ph{prompt}

\smallskip
{\ttfamily\small CANDIDATE COMPLETION:} \ph{completion}
\end{promptbox}

\subsection{Quality metrics per dataset}
\label{app:quality-per-dataset}

\begin{figure*}[t]
\centering
\includegraphics[width=0.62\textwidth, trim=0 324bp 470bp 14bp, clip]{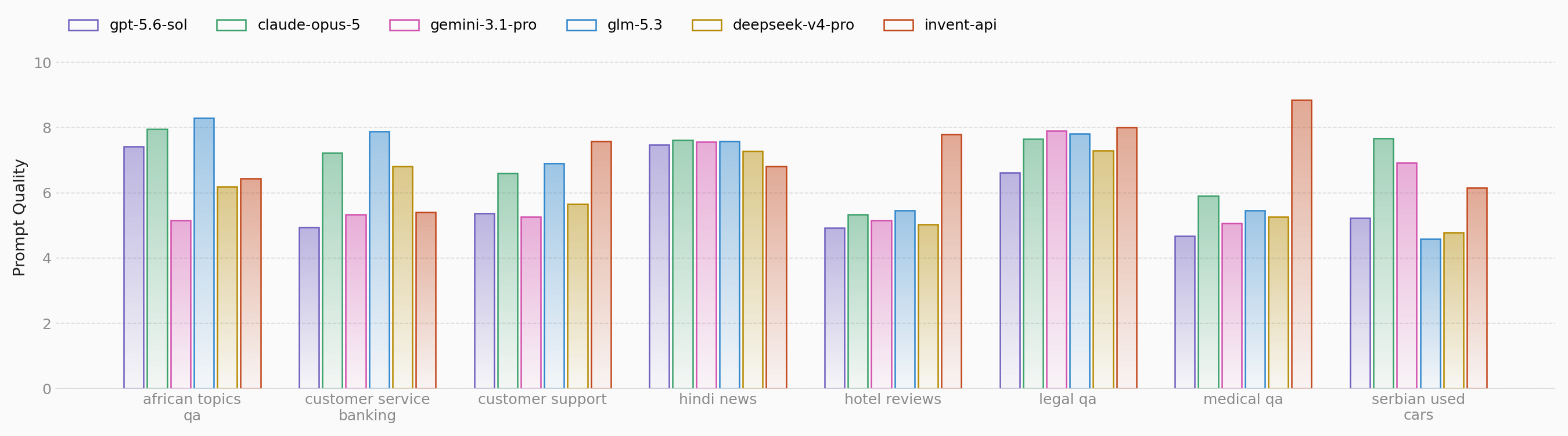}\\[2pt]

\begin{subfigure}[t]{0.49\textwidth}
  \includegraphics[width=\linewidth, trim=0 0 0 36bp, clip]{figures/per_dataset/constrained/pq_bar_per_dataset_200.png}
  \caption{Prompt quality, $N{=}200$}
\end{subfigure}\hfill
\begin{subfigure}[t]{0.49\textwidth}
  \includegraphics[width=\linewidth, trim=0 0 0 36bp, clip]{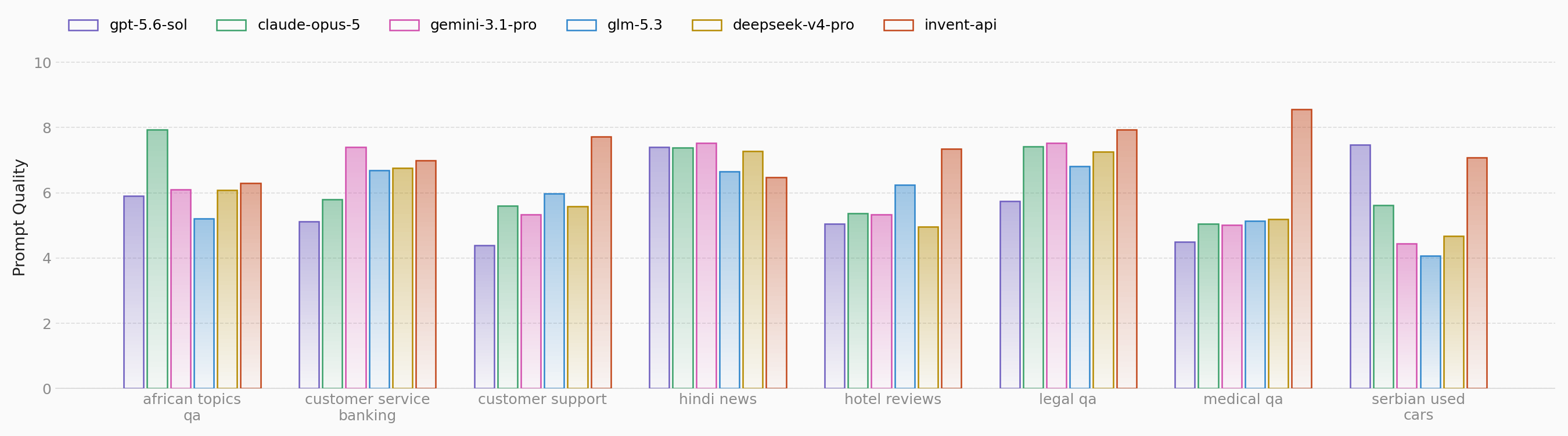}
  \caption{Prompt quality, $N{=}2000$}
\end{subfigure}\\[2pt]
\begin{subfigure}[t]{0.49\textwidth}
  \includegraphics[width=\linewidth, trim=0 0 0 36bp, clip]{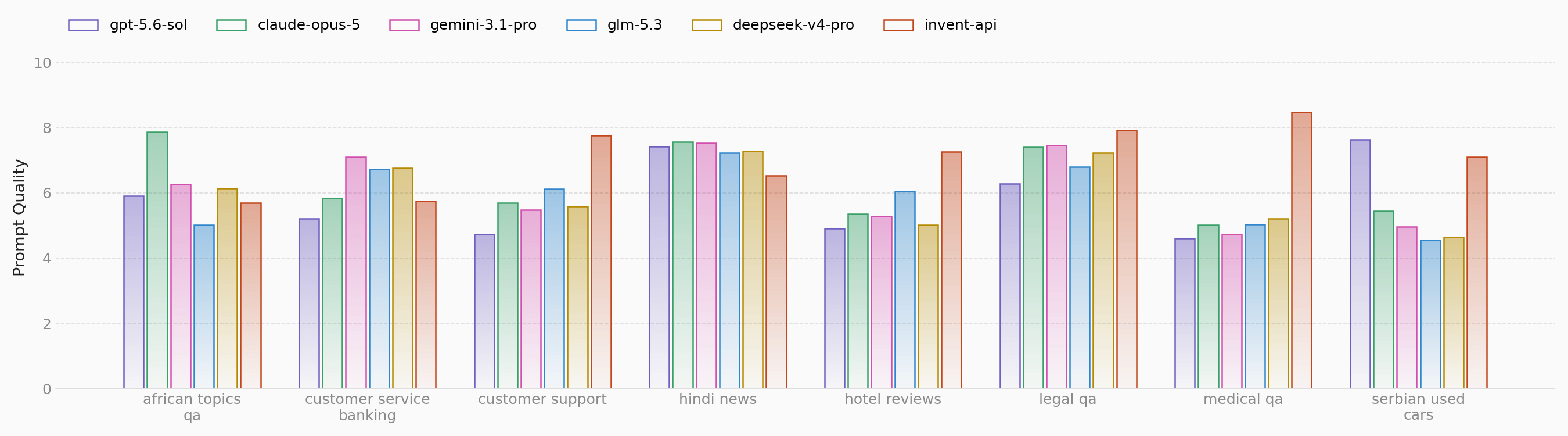}
  \caption{Prompt quality, $N{=}5000$}
\end{subfigure}\hfill
\begin{subfigure}[t]{0.49\textwidth}
  \includegraphics[width=\linewidth, trim=0 0 0 36bp, clip]{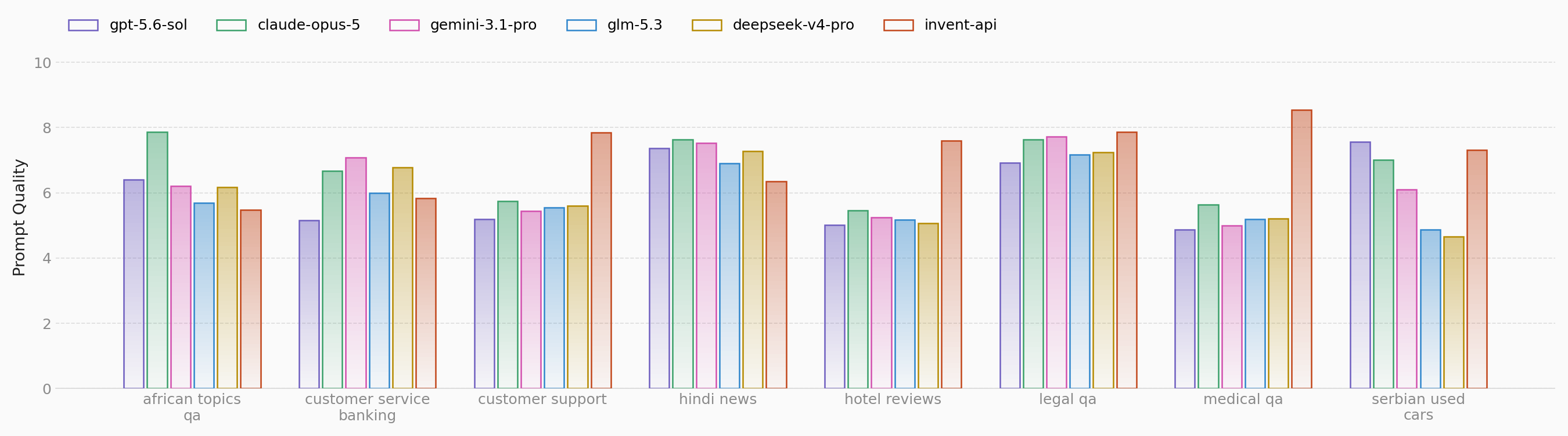}
  \caption{Prompt quality, $N{=}20000$}
\end{subfigure}\\[4pt]

\begin{subfigure}[t]{0.49\textwidth}
  \includegraphics[width=\linewidth, trim=0 0 0 36bp, clip]{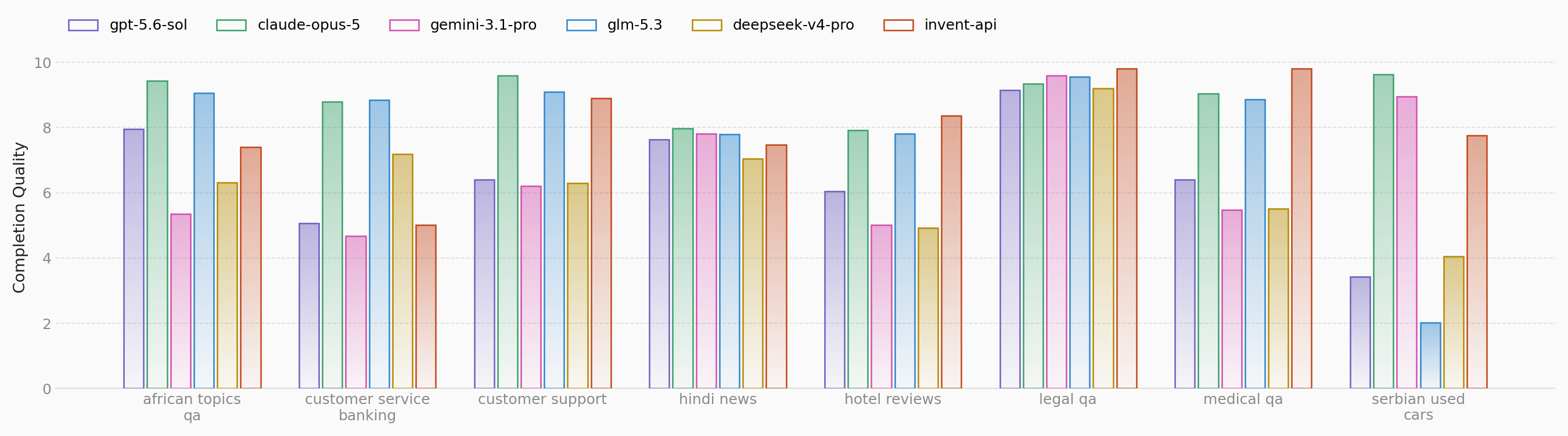}
  \caption{Completion quality, $N{=}200$}
\end{subfigure}\hfill
\begin{subfigure}[t]{0.49\textwidth}
  \includegraphics[width=\linewidth, trim=0 0 0 36bp, clip]{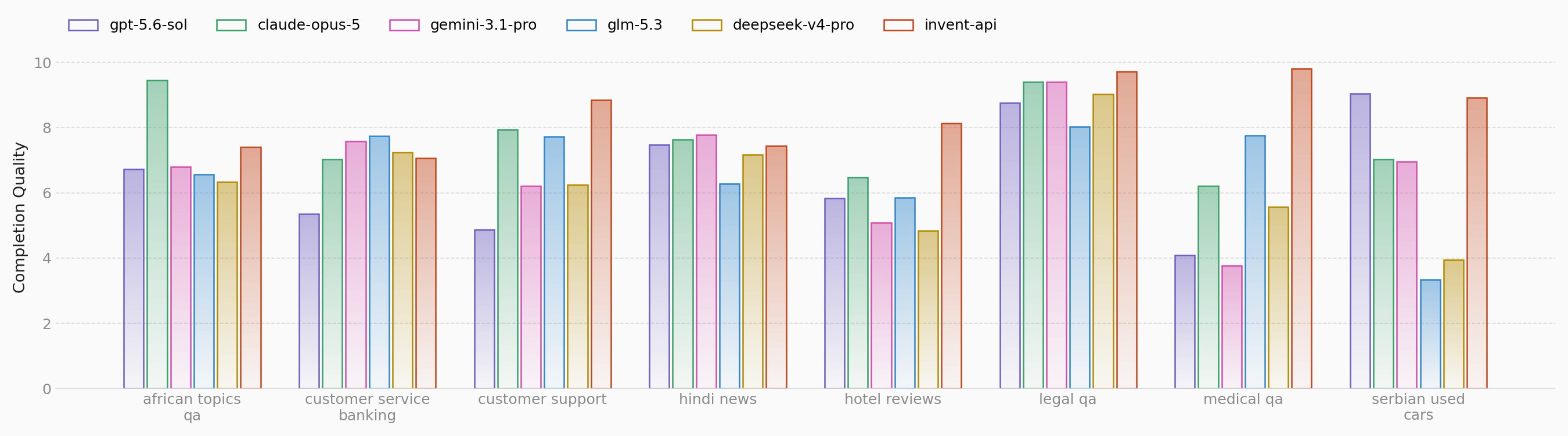}
  \caption{Completion quality, $N{=}2000$}
\end{subfigure}\\[2pt]
\begin{subfigure}[t]{0.49\textwidth}
  \includegraphics[width=\linewidth, trim=0 0 0 36bp, clip]{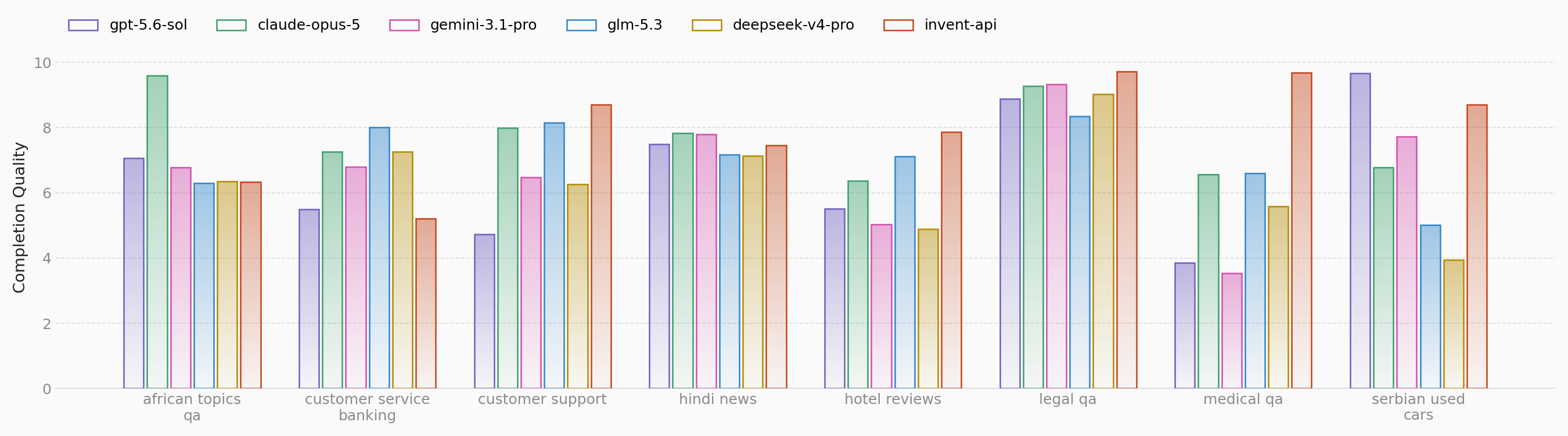}
  \caption{Completion quality, $N{=}5000$}
\end{subfigure}\hfill
\begin{subfigure}[t]{0.49\textwidth}
  \includegraphics[width=\linewidth, trim=0 0 0 36bp, clip]{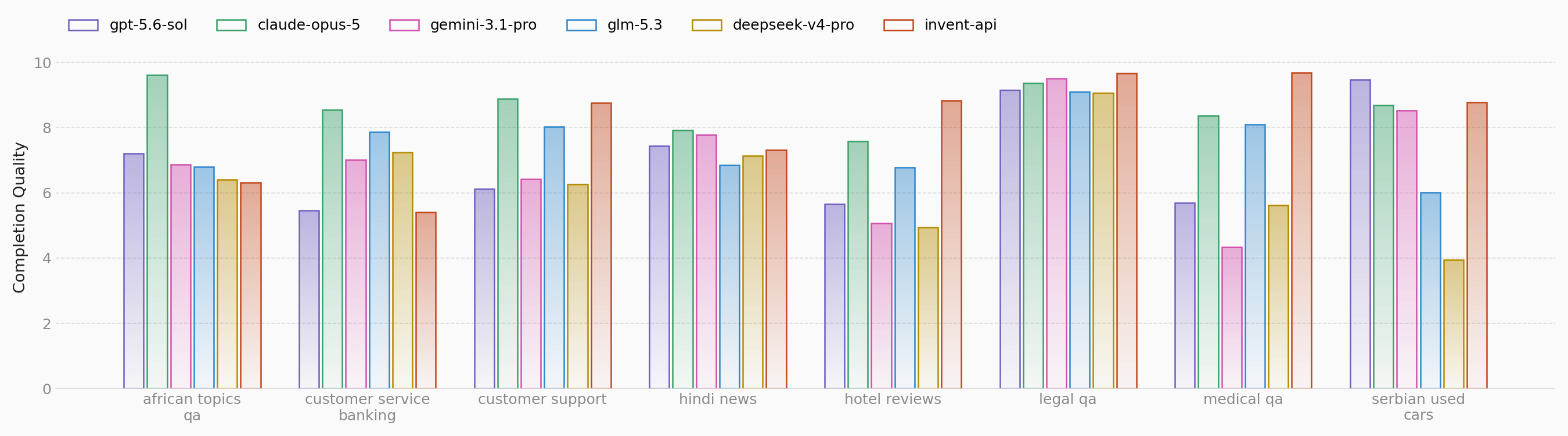}
  \caption{Completion quality, $N{=}20000$}
\end{subfigure}\\[4pt]

\begin{subfigure}[t]{0.49\textwidth}
  \includegraphics[width=\linewidth, trim=0 0 0 36bp, clip]{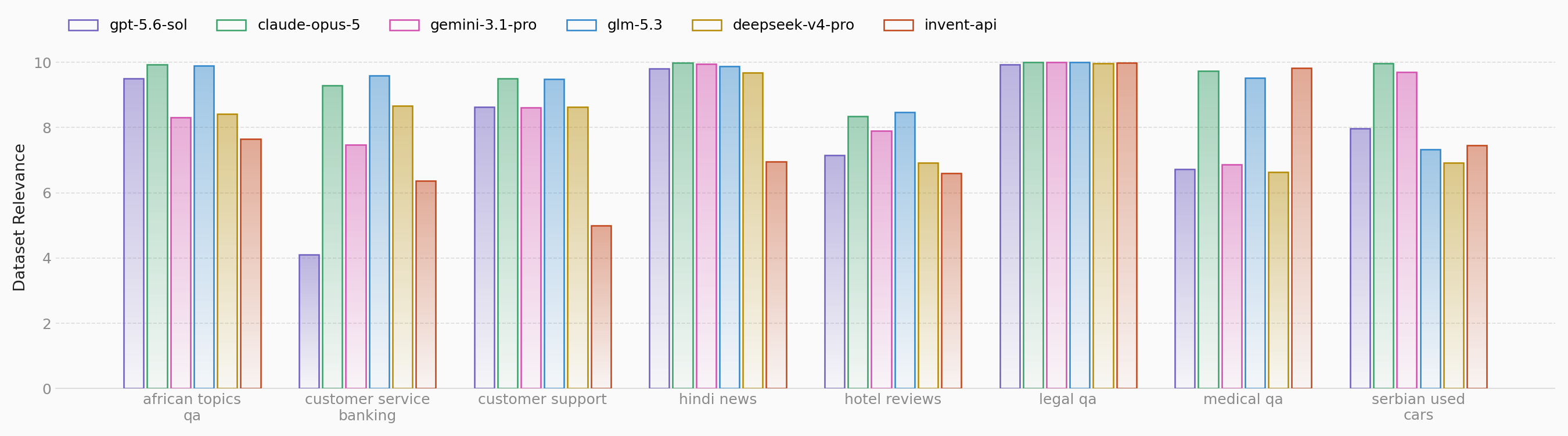}
  \caption{Dataset relevance, $N{=}200$}
\end{subfigure}\hfill
\begin{subfigure}[t]{0.49\textwidth}
  \includegraphics[width=\linewidth, trim=0 0 0 36bp, clip]{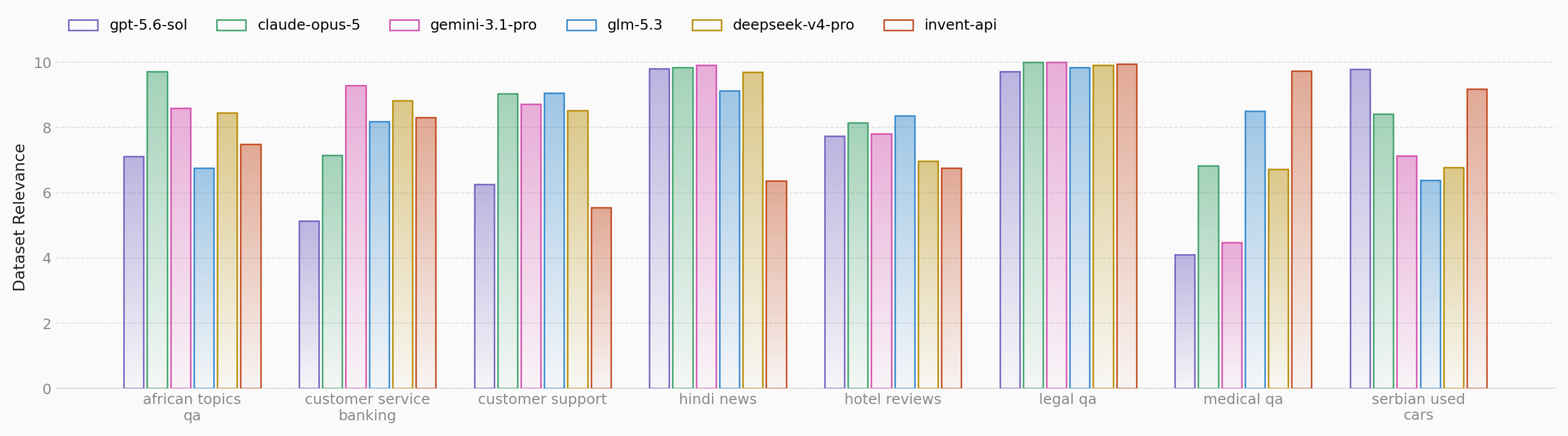}
  \caption{Dataset relevance, $N{=}2000$}
\end{subfigure}\\[2pt]
\begin{subfigure}[t]{0.49\textwidth}
  \includegraphics[width=\linewidth, trim=0 0 0 36bp, clip]{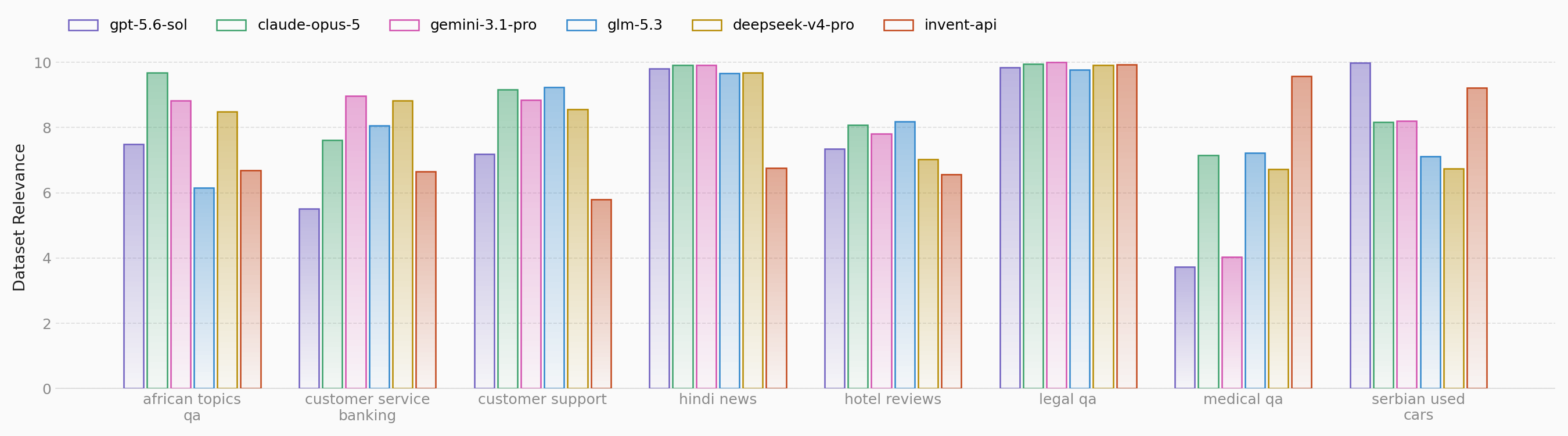}
  \caption{Dataset relevance, $N{=}5000$}
\end{subfigure}\hfill
\begin{subfigure}[t]{0.49\textwidth}
  \includegraphics[width=\linewidth, trim=0 0 0 36bp, clip]{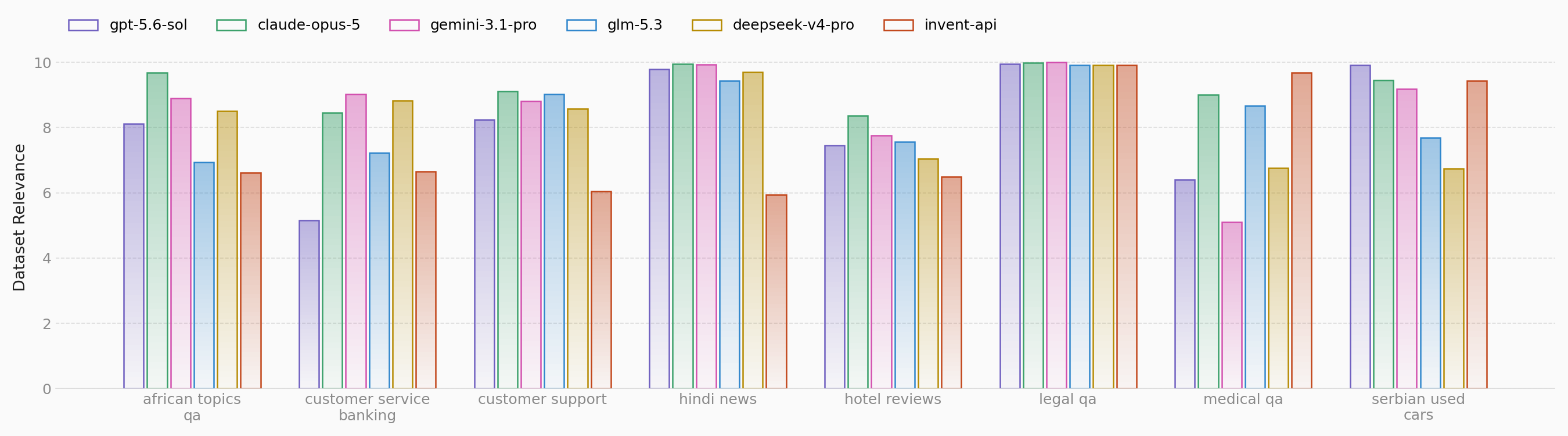}
  \caption{Dataset relevance, $N{=}20000$}
\end{subfigure}

\caption{Per constrained dataset prompt quality, completion quality, and dataset relevance scores for each generator across dataset sizes $N \in \{200, 2000, 5000, 20000\}$.}
\label{fig:constrained_per_dataset_all}
\end{figure*}

\begin{figure*}[t]
\centering
\includegraphics[width=0.62\textwidth, trim=0 324bp 470bp 14bp, clip]{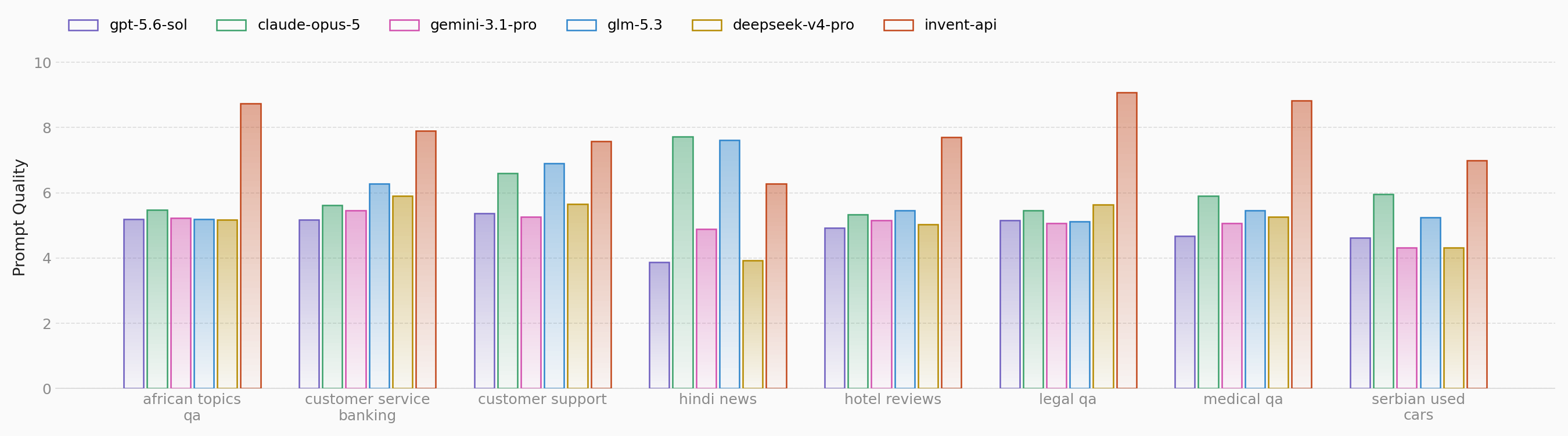}\\[2pt]

\begin{subfigure}[t]{0.49\textwidth}
  \includegraphics[width=\linewidth, trim=0 0 0 36bp, clip]{figures/per_dataset/unconstrained/pq_bar_per_dataset_200_uc.png}
  \caption{Prompt quality, $N{=}200$}
\end{subfigure}\hfill
\begin{subfigure}[t]{0.49\textwidth}
  \includegraphics[width=\linewidth, trim=0 0 0 36bp, clip]{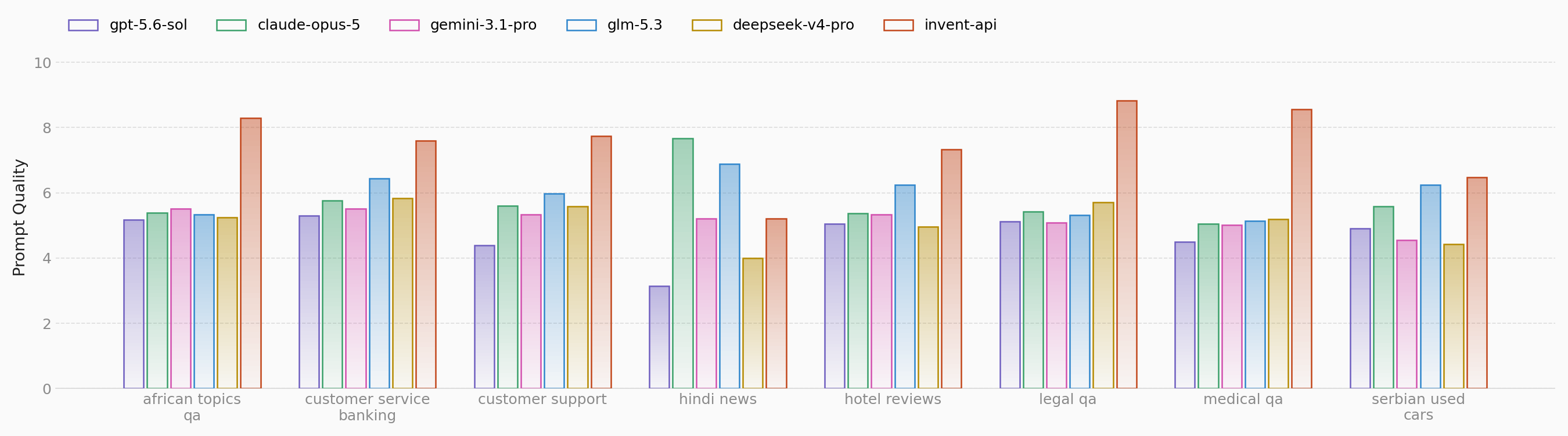}
  \caption{Prompt quality, $N{=}2000$}
\end{subfigure}\\[2pt]
\begin{subfigure}[t]{0.49\textwidth}
  \includegraphics[width=\linewidth, trim=0 0 0 36bp, clip]{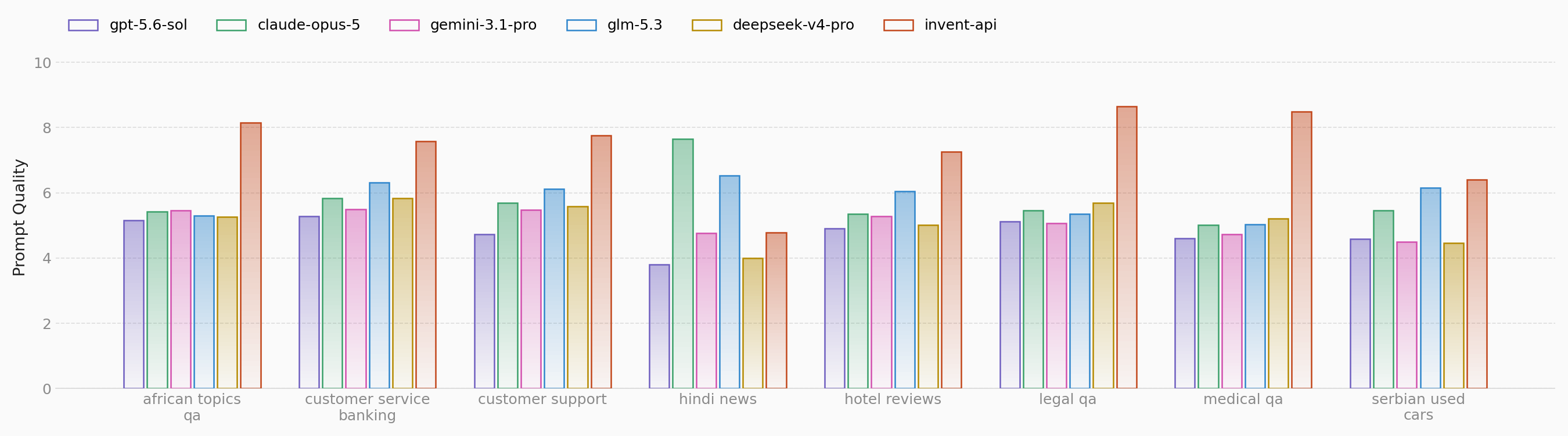}
  \caption{Prompt quality, $N{=}5000$}
\end{subfigure}\hfill
\begin{subfigure}[t]{0.49\textwidth}
  \includegraphics[width=\linewidth, trim=0 0 0 36bp, clip]{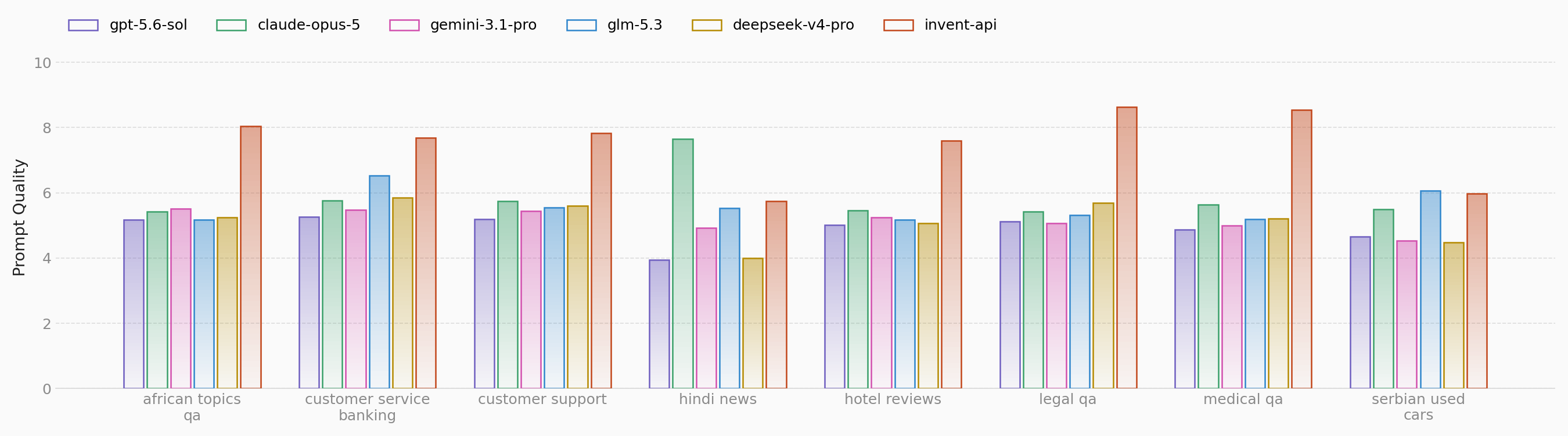}
  \caption{Prompt quality, $N{=}20000$}
\end{subfigure}\\[4pt]

\begin{subfigure}[t]{0.49\textwidth}
  \includegraphics[width=\linewidth, trim=0 0 0 36bp, clip]{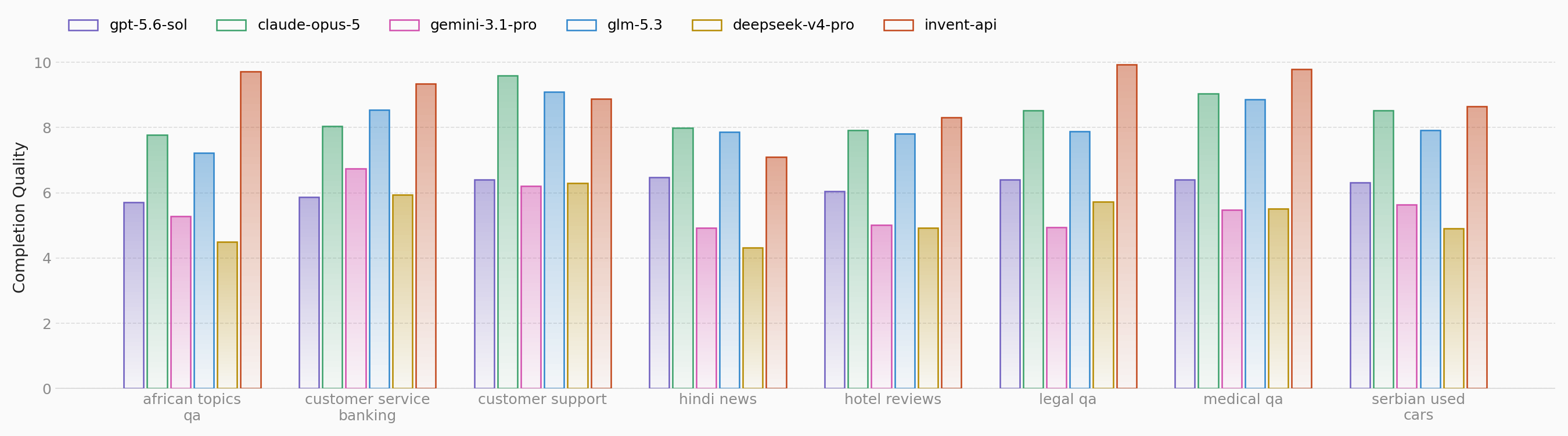}
  \caption{Completion quality, $N{=}200$}
\end{subfigure}\hfill
\begin{subfigure}[t]{0.49\textwidth}
  \includegraphics[width=\linewidth, trim=0 0 0 36bp, clip]{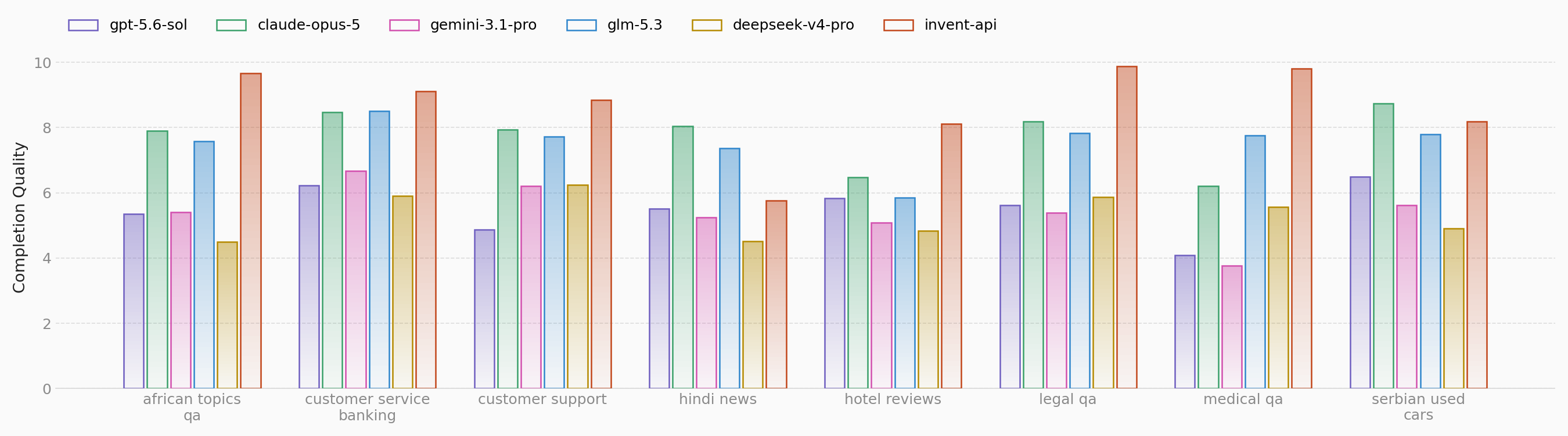}
  \caption{Completion quality, $N{=}2000$}
\end{subfigure}\\[2pt]
\begin{subfigure}[t]{0.49\textwidth}
  \includegraphics[width=\linewidth, trim=0 0 0 36bp, clip]{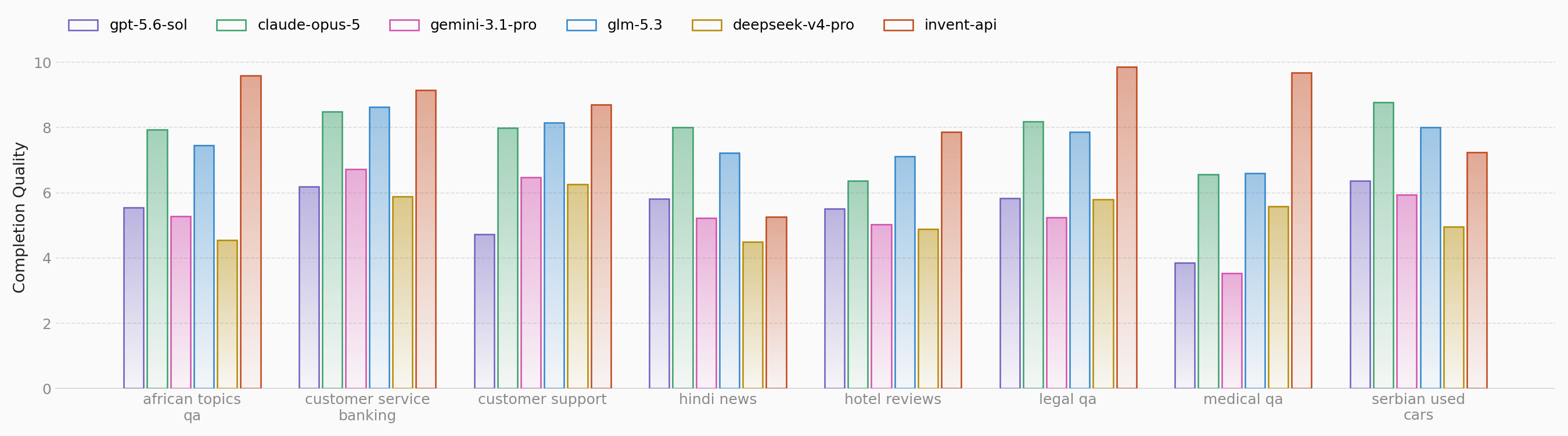}
  \caption{Completion quality, $N{=}5000$}
\end{subfigure}\hfill
\begin{subfigure}[t]{0.49\textwidth}
  \includegraphics[width=\linewidth, trim=0 0 0 36bp, clip]{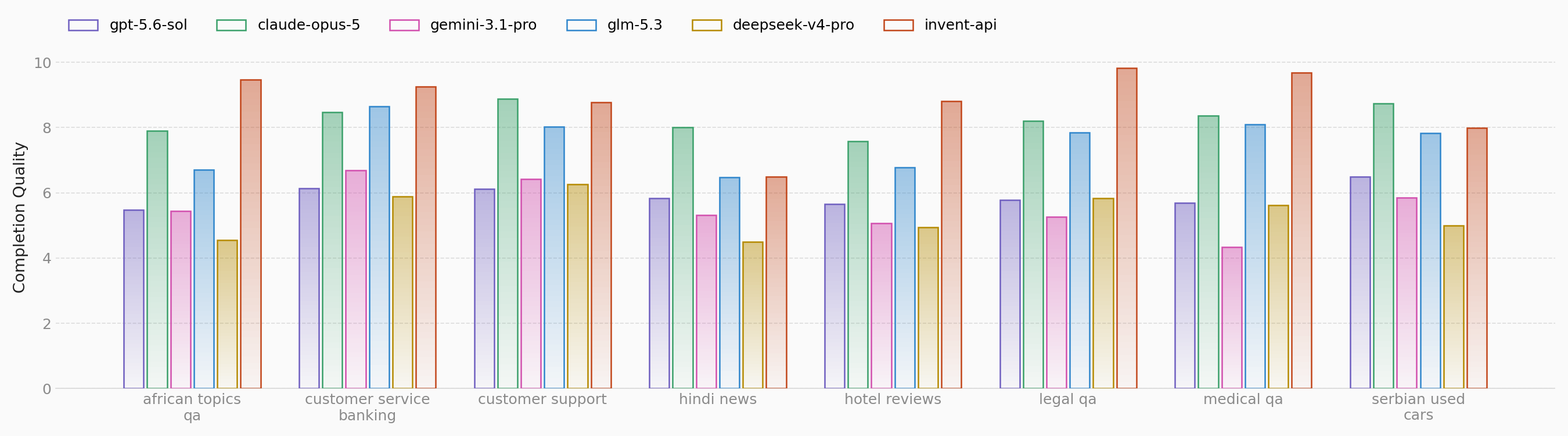}
  \caption{Completion quality, $N{=}20000$}
\end{subfigure}\\[4pt]

\begin{subfigure}[t]{0.49\textwidth}
  \includegraphics[width=\linewidth, trim=0 0 0 36bp, clip]{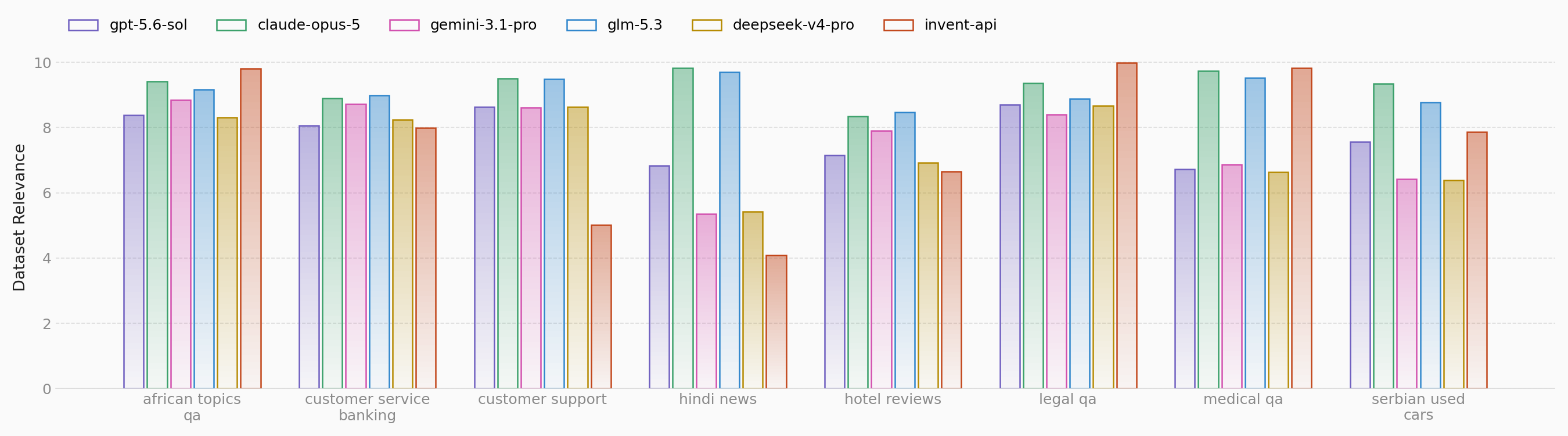}
  \caption{Dataset relevance, $N{=}200$}
\end{subfigure}\hfill
\begin{subfigure}[t]{0.49\textwidth}
  \includegraphics[width=\linewidth, trim=0 0 0 36bp, clip]{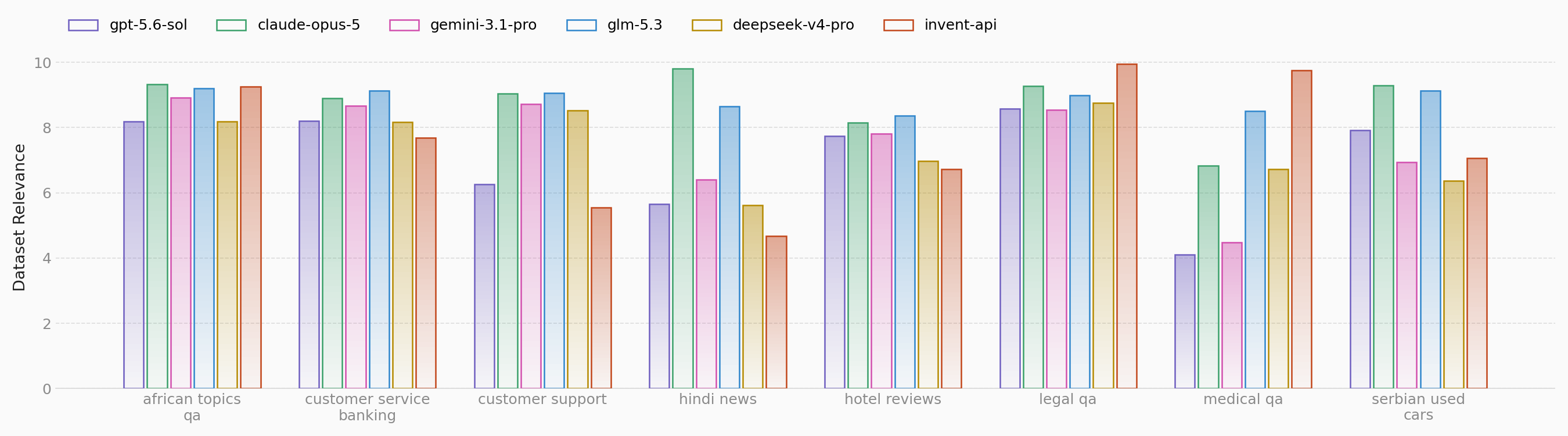}
  \caption{Dataset relevance, $N{=}2000$}
\end{subfigure}\\[2pt]
\begin{subfigure}[t]{0.49\textwidth}
  \includegraphics[width=\linewidth, trim=0 0 0 36bp, clip]{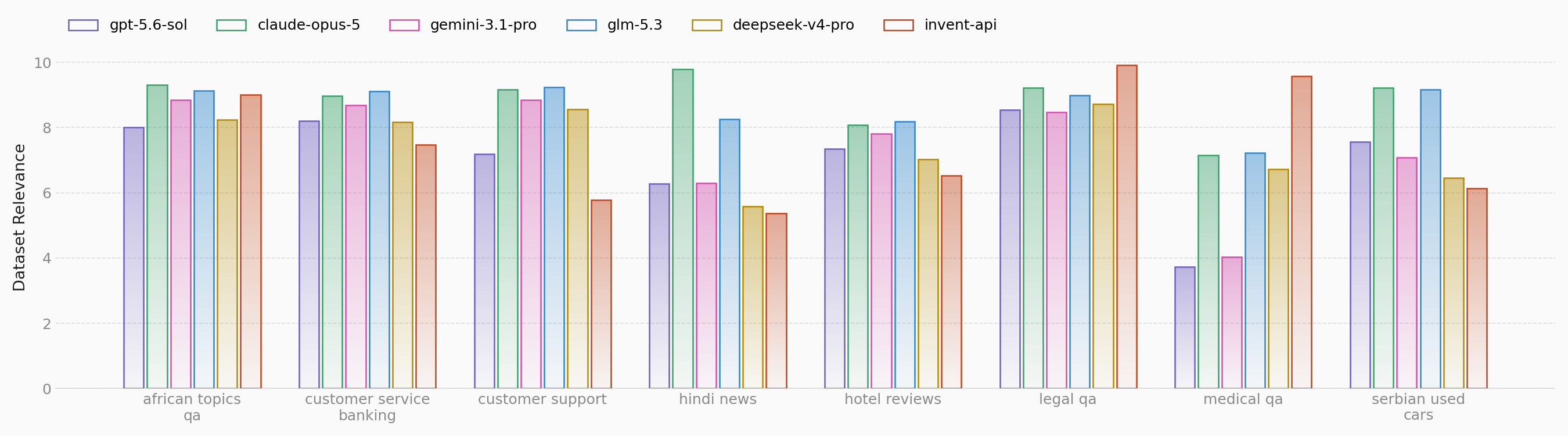}
  \caption{Dataset relevance, $N{=}5000$}
\end{subfigure}\hfill
\begin{subfigure}[t]{0.49\textwidth}
  \includegraphics[width=\linewidth, trim=0 0 0 36bp, clip]{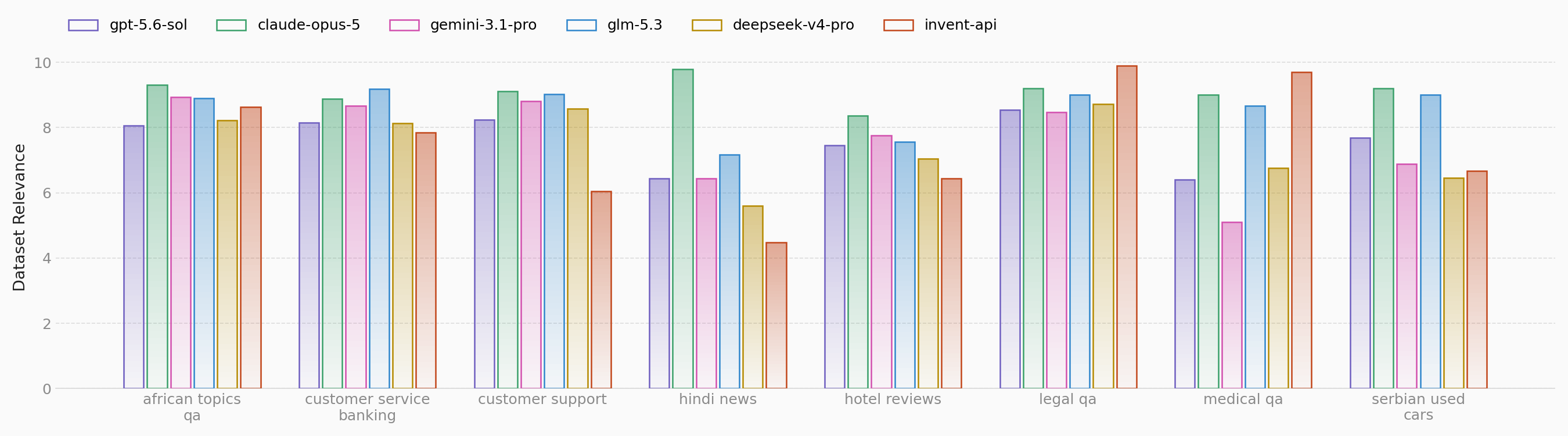}
  \caption{Dataset relevance, $N{=}20000$}
\end{subfigure}

\caption{Per unconstrained dataset prompt quality, completion quality, and dataset relevance scores for each generator across dataset sizes $N \in \{200, 2000, 5000, 20000\}$.}
\label{fig:unconstrained_per_dataset_all}
\end{figure*}

Figures~\ref{fig:constrained_per_dataset_all} and~\ref{fig:unconstrained_per_dataset_all} report per-dataset quality results at all 4 scales for the constrained and unconstrained settings, respectively.

\subsection{Dataset Generation Queries}
Table~\ref{tab:queries} reports the uncontrained and constrained queries used to generate other datasets in our evaluation set. 
\label{app:dataset-generation-queries}

\begin{table}[h!]
\centering
\footnotesize
\setlength{\extrarowheight}{2pt}
\begin{tabularx}{\textwidth}{@{}%
  >{\RaggedRight\arraybackslash}p{1.7cm}%
  >{\RaggedRight\arraybackslash}X%
  >{\RaggedRight\arraybackslash}X@{}}
\toprule
\textbf{Dataset} & \textbf{Unconstrained Query} & \textbf{Constrained Query} \\
\midrule

Customer Support Banking &
Dataset of customer service responses guiding users through credit card activation, blocking, and mortgage inquiries. &
Dataset of customer service conversations guiding frustrated users through credit card activation, blocking, and mortgage inquiries. The customer service agent is always welcoming and polite. \\
\midrule

Hindi News &
Dataset containing Hindi news articles sourced from various Indian news websites, paired with their corresponding headlines and summaries. &
Dataset containing Hindi news articles sourced from various Indian news websites about sports, paired with their corresponding summaries. The prompt is always the article headline, and the completion is the article summary. \\
\midrule

Customer Support General &
A dataset of customer support inquiries and corresponding responses, covering issues such as product information, technical troubleshooting, billing disputes, and service returns. &
 \\
\midrule

Medical QA &
Dataset of medical question--answer pairs covering diagnoses, treatments, drug explanations, and healthcare roles. Each entry includes a prompt and a detailed, expert-generated response. &
 \\
\midrule

Hotel Reviews &
A collection of customer reviews describing hotel stays, focusing on service, room quality, location, and amenities. The dataset should include both positive and negative experiences. &
 \\
\bottomrule
\end{tabularx}
\caption{The single request queries used to generate datasets. Queries are choosen to represent a range
of 1) task types, 2) languages and 3) domains. We evaluate on both \textbf{unconstrained} and \textbf{constrained} versions of the query, where the constrained query requires additional requirements such as  1) tone constraints, 2) json format, 3) multiple choice output restrictions. 
}
\label{tab:queries}
\end{table}

\subsection{Output Language Consistency}
\label{langid-consistency}
Table~\ref{tab:hindi_news} reports output language consistency on \texttt{hindi\_news} (2000 samples from a pool of 20000). We count any output not detected as Hindi as a language error, including outputs in Marathi and Nepali. \inventapi was the most consistent system, with only 3 non-Hindi outputs (0.15\%), ahead of Claude Opus~5 (7; 0.35\%), GLM-5.3 (49; 2.45\%), GPT-5.6-sol (50; 2.5\%), Gemini~3.1~Pro (77; 3.85\%), and DeepSeek~V4~Pro (145; 7.25\%). Marathi was the most common error for every system except GLM-5.3, whose errors were mostly English (38 of 49). DeepSeek~V4~Pro's error rate was driven almost entirely by Marathi (144 outputs). The only other languages observed were Nepali (DeepSeek~V4~Pro and Gemini~3.1~Pro, three outputs combined) and a single Lithuanian output from \inventapi.

\begin{table}[ht]
\centering
\footnotesize
\setlength{\extrarowheight}{2pt}
\begin{tabularx}{\linewidth}{@{}l r r >{\RaggedRight\arraybackslash}X@{}}
\toprule
\textbf{Model} & \textbf{Not in language} & \textbf{\%} & \textbf{Detected languages} \\
\midrule
claude-opus-5        & 7/2000   & 0.35 & hi 1993, mr 7 \\
deepseek-v4-pro-0813 & 145/2000 & 7.25 & hi 1855, mr 144, ne 1 \\
gemini-3.1-pro       & 77/2000  & 3.85 & hi 1923, mr 75, ne 2 \\
gpt-5.6-sol          & 50/2000  & 2.50 & hi 1950, mr 50 \\
glm-5.3              & 49/2000  & 2.45 & hi 1951, en 38, mr 11 \\
\rowcolor{gray!15}
\textbf{invent-api}  & \textbf{3/2000}   & \textbf{0.15} & \textbf{hi 1997, mr 2, lt 1} \\
\bottomrule
\end{tabularx}
\caption{Output language consistency on \texttt{hindi\_news} (2k samples drawn from 20k). \emph{Not in language} counts all outputs detected as any language other than Hindi, including Marathi and Nepali. Language codes: hi = Hindi, mr = Marathi, ne = Nepali, en = English, lt = Lithuanian.}
\label{tab:hindi_news}
\end{table}

\subsection{Lexical diversity: constrained setting}
\label{app:lexical_constrained}

Table~\ref{tab:app-lexical-constrained} reports the same three lexical prompt metrics in the
constrained setting, at the $2{,}000$, $5{,}000$, and $20{,}000$-sample scales (each on a fixed
$2{,}000$ prompt random sample). The pattern mirrors the unconstrained results in
Table~\ref{tab:lexical-prompt}. The \inventapi has the least redundant prompts (lowest compression ratio at every size) and a near-zero duplicate rate at every scale, while baseline duplication grows with dataset
size. On raw $n$-gram diversity the \inventapi and GLM-5.3 are close and alternate at the top (GLM-5.3
edges ahead at $5{,}000$). Overall, constraining generation lowers absolute duplication rate for all baseline generators whereas \inventapi saw a minor increase from 0.0\% (in unconstrained setting) to 0.1\% (in constrained setting). 

\begin{table*}[h!]
\centering
\small
\setlength{\tabcolsep}{5pt}
\begin{tabular}{lccccccccc}
\toprule
& \multicolumn{3}{c}{NGD$_{\text{prompt}}\uparrow$} & \multicolumn{3}{c}{CR$_{\text{prompt}}\downarrow$} & \multicolumn{3}{c}{Dup$_{\text{prompt}}$ (\%)$\downarrow$} \\
\cmidrule(lr){2-4}\cmidrule(lr){5-7}\cmidrule(lr){8-10}
Model & 2K & 5K & 20K & 2K & 5K & 20K & 2K & 5K & 20K \\
\midrule
\inventapi & \textbf{1.69} & 1.69 & \textbf{1.72} & \textbf{3.55} & \textbf{3.53} & \textbf{3.50} & \textbf{0.1} & \textbf{0.1} & \textbf{0.1} \\
GLM-5.3 & 1.65 & \textbf{1.80} & 1.71 & 4.59 & 3.75 & 3.83 & 0.9 & 1.2 & 3.9 \\
Gemini-3.1-pro & 1.34 & 1.39 & 1.53 & 5.18 & 4.75 & 4.03 & 2.2 & 2.3 & 1.8 \\
GPT-5.6 Sol & 1.56 & 1.57 & 1.53 & 4.16 & 4.22 & 4.29 & 2.2 & 3.7 & 5.0 \\
Claude Opus~5 & 1.61 & 1.56 & 1.46 & 4.00 & 4.02 & 4.09 & 2.2 & 3.9 & 4.3 \\
DeepSeek-V4-Pro & 1.12 & 1.13 & 1.12 & 5.15 & 5.12 & 5.17 & 11.5 & 11.0 & 11.6 \\
\bottomrule
\end{tabular}
\caption{Lexical diversity and duplication in the \emph{constrained} setting, averaged over
the eight datasets (computed for each dataset size on a fixed $2{,}000$ prompt random sample). n-gram diversity, NGD ($\uparrow$ is better), Compression Ratio, CR ($\downarrow$ is better), Duplicate Rate (Dup \%, $\downarrow$ is better). The best per column is highlighted in bold.}
\label{tab:app-lexical-constrained}
\end{table*}

\subsection{Post-training results}
\label{app:qwen-ranking}

\begin{figure}[h!]
    \centering
    \includegraphics[width=0.5\linewidth]{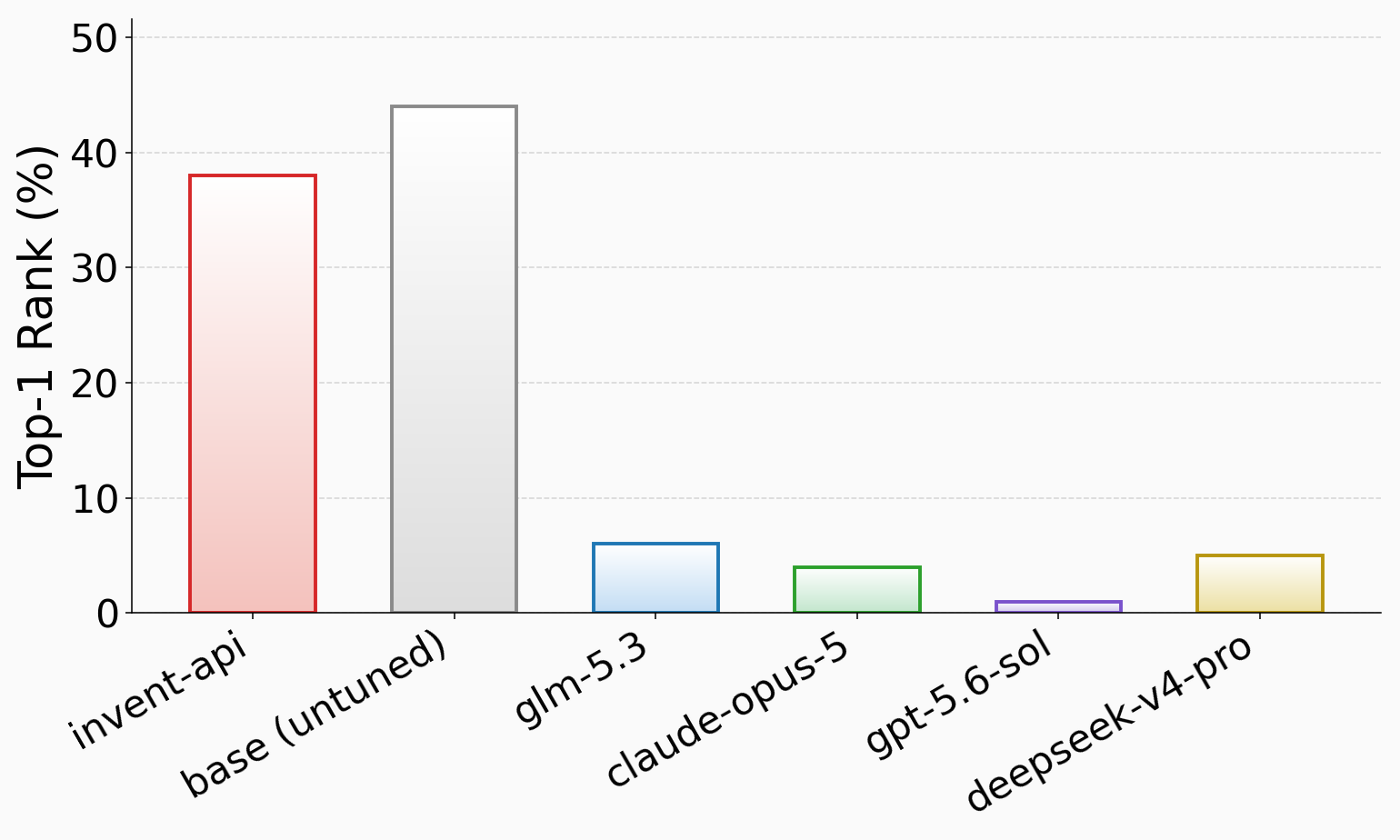}
    \caption{\textbf{Top-1 rank share for Qwen3.5-9B fine-tunes.} Percentage of held-out medical prompts on which each model ranks first (all six responses judged together per prompt); grey is the untuned base. Qwen3.5-9B has an unusually strong base that edges the \inventapi fine-tune on both first-place rate ($44\%$ vs.\ $38\%$) and mean rank ($2.17$ vs.\ $2.28$). \inventapi fine-tuned variant nonetheless remains far ahead of all variants corresponding to other generators.}
    \label{fig:qwen-ranking}
\end{figure}

Apart from Llama-3.3-70B and Gemma-4-31B-it, we also fine-tune Qwen3.5-9B. As shown in Figure~\ref{fig:qwen-ranking}, Qwen3.5-9B is an really strong starting point, edging \inventapi fine-tuned variant on both first-place rate (44\% vs.\ 38\%) and mean rank (2.17 vs.\ 2.28). \inventapi fine-tune nonetheless remains ahead of every variant corresponding to the other generators. 

Across all three chosen model architectures, \inventapi is consistently the top-ranked generator relative to the fine-tunes from every other generator. We note that these results are obtained with a modest 20K fine-tuning set. Prior work shows that supervised fine-tuning performance scales predictably with the amount of training data~\citep{zhang2024scalingmeetsllmfinetuning}, so we expect the margin of the \inventapi fine-tune over the strongest base models to widen further at larger data budgets.

\subsection{Dataset Diversity vs Dataset Quality Across Different Scales}
\label{app:diversity_all_plots}
\begin{figure}[h!]
    \centering
    \includegraphics[width=0.5\linewidth]{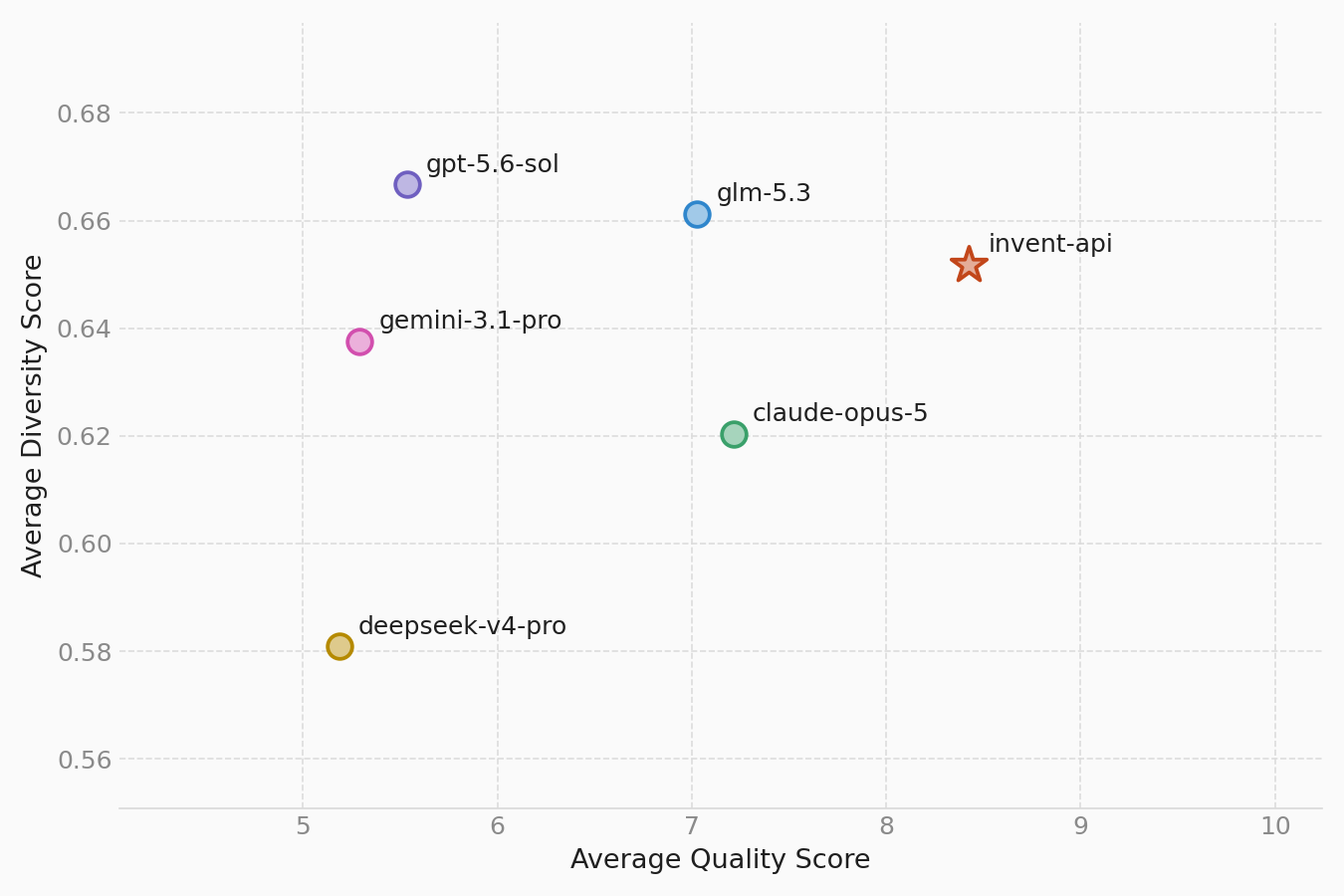}
    \caption{\textbf{Diversity vs.\ quality on 200-sample datasets.} Each point shows the average quality score (x-axis) and average diversity score, measured by DCScore ($\tau = 0.1$) (y-axis), across 8 dataset in unconstrained setting.}
    \label{fig:dq_200_app}
\end{figure}

\begin{figure}[h!]
    \centering
    \includegraphics[width=0.5\linewidth]{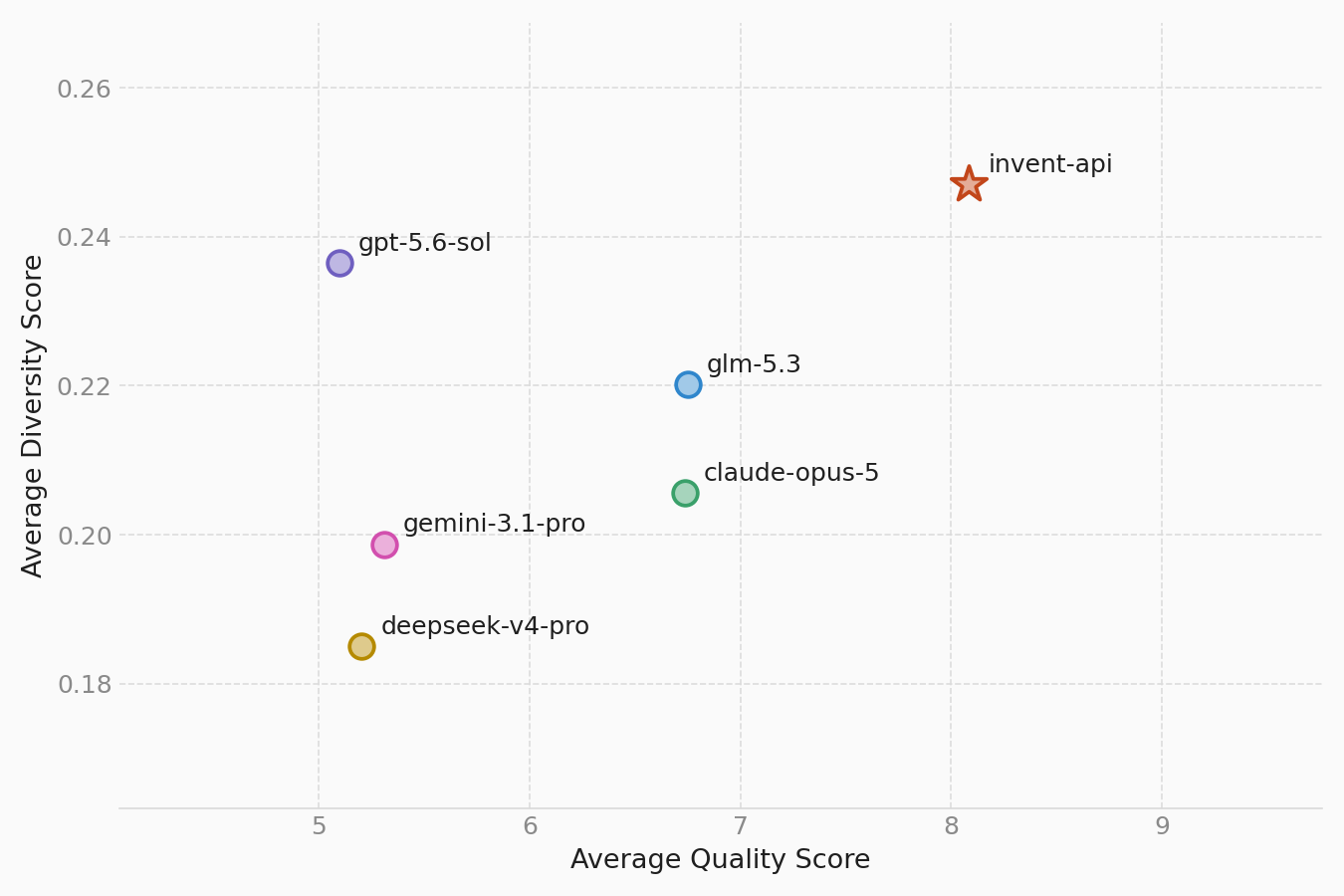}
    \caption{\textbf{Diversity vs.\ quality on 2000-sample datasets.} Each point shows the average quality score (x-axis) and average diversity score, measured by DCScore ($\tau = 0.1$) (y-axis), across 8 dataset in unconstrained setting.}
    \label{fig:dq_2000_app}
\end{figure}

\begin{figure}[h!]
    \centering
    \includegraphics[width=0.5\linewidth]{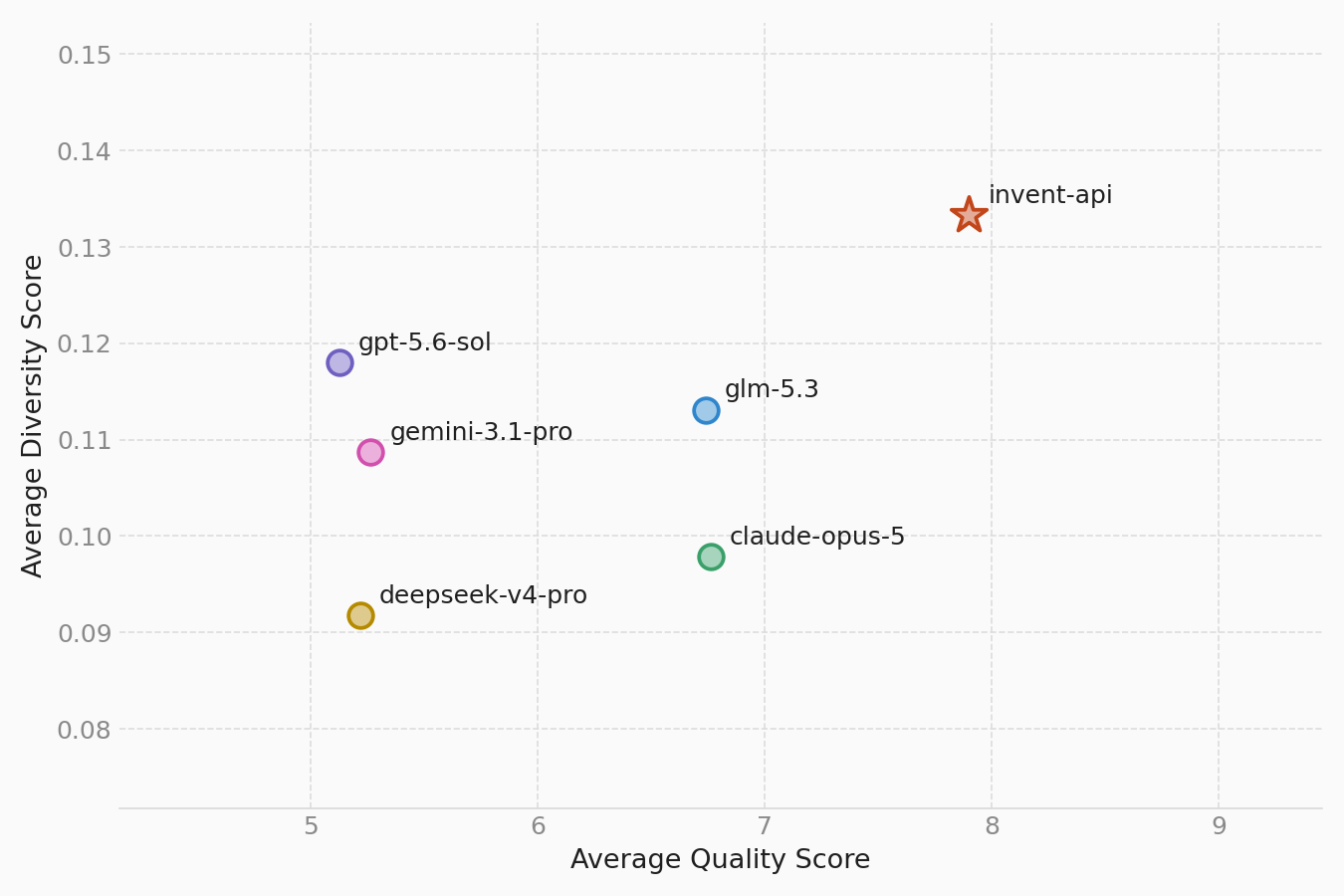}
    \caption{\textbf{Diversity vs.\ quality on 5000-sample datasets.} Each point shows the average quality score (x-axis) and average diversity score, measured by DCScore ($\tau = 0.1$) (y-axis), across 8 dataset in unconstrained setting.}
    \label{fig:dq_5000_app}
\end{figure}

\begin{figure}[h!]
    \centering
    \includegraphics[width=0.5\linewidth]{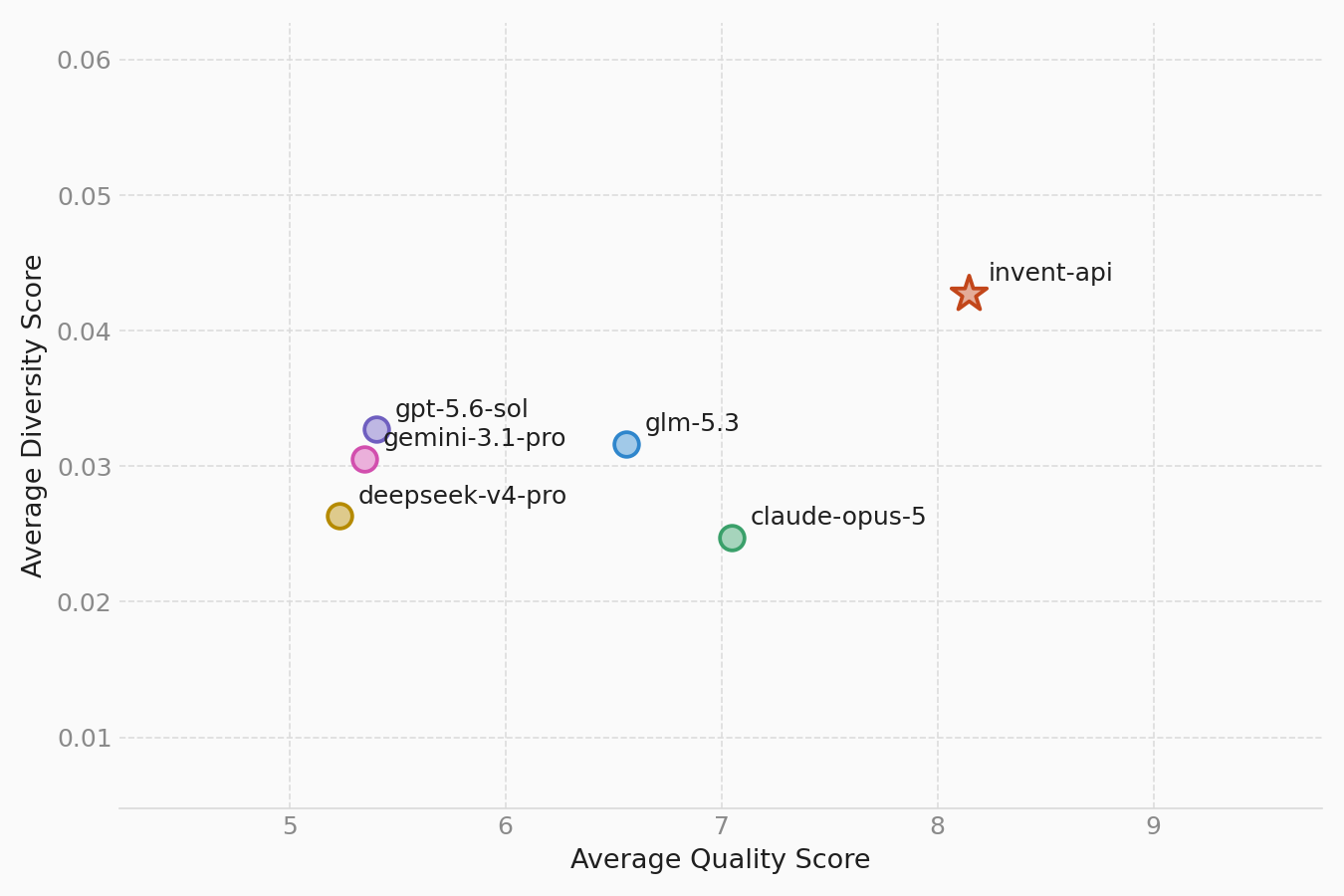}
    \caption{\textbf{Diversity vs.\ quality on 20000-sample datasets.} Each point shows the average quality score (x-axis) and average diversity score, measured by DCScore ($\tau = 0.1$) (y-axis), across 8 dataset in unconstrained setting.}
    \label{fig:dq_20000_app}
\end{figure}

Figures~\ref{fig:dq_200_app}--\ref{fig:dq_20000_app} plot average diversity against average quality for datasets of 200, 2{,}000, 5{,}000, and 20{,}000 samples: \texttt{invent-api} achieves the highest quality at every scale and the highest diversity at every scale except 200 samples, where all APIs are on par.

\FloatBarrier

\subsection{Ranking Evaluation}
\label{app:ranking_evaluation}
This rubric compares models directly rather than scoring them one at a time. The models fine-tuned on the generated dataset and the un-tuned base model each answer the same held-out set of medical questions. For each question, the judge sees all responses in a single call and returns a best-to-worst ordering. 

\newtcolorbox{titlecard1}[1]{
  enhanced, breakable,
  colback=cardbg, colframe=cardframe, boxrule=0.5pt, arc=4pt,
  coltitle=white, fonttitle=\bfseries,
  colbacktitle=titleblue,
  attach boxed title to top left={xshift=10pt, yshift=-\tcboxedtitleheight/2},
  boxed title style={colframe=titleblue, colback=titleblue, arc=3pt,
                     boxrule=0pt, top=2pt, bottom=2pt, left=6pt, right=6pt},
  title={#1},
  top=12pt, left=9pt, right=9pt, bottom=9pt
}

\newtcolorbox{promptbox1}[1]{
  enhanced, breakable,
  colback=white, colframe=sectionframe, boxrule=0.5pt, arc=3pt,
  colbacktitle=sectionbg, coltitle=black, fonttitle=\bfseries,
  title={#1},
  left=9pt, right=9pt, top=7pt, bottom=7pt,
  toptitle=3pt, bottomtitle=3pt, lefttitle=9pt,
  before skip=8pt, after skip=8pt
}

\setlist[scale]{leftmargin=1.4em, itemsep=3pt, topsep=3pt, parsep=0pt, label={}}

\setlength{\parindent}{0pt}
\setlength{\parskip}{4pt}

\begin{promptbox1}{Ranking Prompt}
\textbf{Inputs}\\[2pt]
Below you are given a prompt from a user for an AI assistant, enclosed between
\texttt{<prompt>} and \texttt{</prompt>}, and a set of responses from an AI assistant,
enclosed between \texttt{<responses>} and \texttt{</responses>}.

\texttt{<prompt>\{PROMPT\}</prompt>}\\
\texttt{<responses>}\\
\texttt{\{RESPONSES\}}\\
\texttt{</responses>}

\medskip
\textbf{Evaluation Rubric}\\[2pt]
Your job is to evaluate and then rank the responses from best to worst. Please act as an
impartial judge and evaluate the quality of the following \texttt{\{N\}} responses to the
prompt. You will output two fields, \enquote{\texttt{ranking}} and
\enquote{\texttt{ranking\_explanation}}. \enquote{\texttt{ranking\_explanation}} explains your
rationale for why a completion is better than the others. \enquote{\texttt{ranking}} is a list
of exactly \texttt{\{N\}} integers corresponding to the indices of the responses (the responses
are zero-indexed), ranked from best to worst.

Evaluate each response based on the following criteria:

\textsc{Accuracy} (Critical):
\begin{itemize}
  \item Does the response provide factually correct information? Any hallucination or error
        should automatically result in a lower rank.
  \item Is every factual claim in the response correct?
  \item Does the response avoid hallucination, fabrication, or misleading implications?
  \item If a response contains any factual error, it must be ranked below all factually
        correct responses, regardless of how well it performs on other criteria.
\end{itemize}

\textsc{Instruction Following} (Critical):
\begin{itemize}
  \item Does the response directly answer the question asked --- no more, no less?
  \item Does it address every explicit and implicit requirement of the prompt?
  \item Does it avoid answering a different or easier version of the question?
  \item Penalty: partial answers, redirections, or responses that technically answer but miss
        the core intent should be ranked lower.
\end{itemize}

\textsc{Completeness vs.\ Conciseness} (Medium Weight):
\begin{itemize}
  \item Is the response as concise as possible without omitting necessary context?
  \item Does added detail genuinely improve usefulness, or is it padding?
  \item Penalty: unnecessary verbosity (adds noise, reduces clarity) and excessive brevity
        (omits context the user would reasonably need) should both be penalized relative to
        each other.
\end{itemize}

\textsc{Clarity \& Naturalness} (Low Weight --- tiebreaker):
\begin{itemize}
  \item Is the response grammatically correct and easy to understand?
  \item Does it read naturally, without awkward phrasing or redundancy?
  \item Use this criterion only to break ties when higher-priority criteria are equal across
        responses.
\end{itemize}

Act as an impartial judge. Do not favor any response based on length or style alone. If all
responses are truly indistinguishable, rank by lowest index first.

\medskip
\textbf{Output Format}\\[2pt]
Your output should be \textsc{a valid JSON object} (not in a code block), encoded in UTF-8,
exactly matching the following JSON schema:
\begin{verbatim}
{
  "additionalProperties": false,
  "properties": {
    "ranking_explanation": {
      "description": "A brief rationale comparing the responses
                      and explaining the ranking order.",
      "title": "Ranking Explanation",
      "type": "string"
    },
    "ranking": {
      "description": "List of zero-indexed response indices
                      ordered from best to worst.",
      "items": { "type": "integer" },
      "title": "Ranking",
      "type": "array"
    }
  },
  "required": ["ranking_explanation", "ranking"],
  "title": "RankingExtractionOutput",
  "type": "object"
}
\end{verbatim}
Your output should contain \textsc{only} the JSON object and nothing else. \textbf{Do not}
include explanations or any other text besides the JSON object.
\end{promptbox1}

\subsection{Selected Dataset Samples}
\label{sec:selected-data-samples}
The subsequent table includes one sample including prompt and completion from each dataset produced by each generator. 

\includepdf[pages=-]{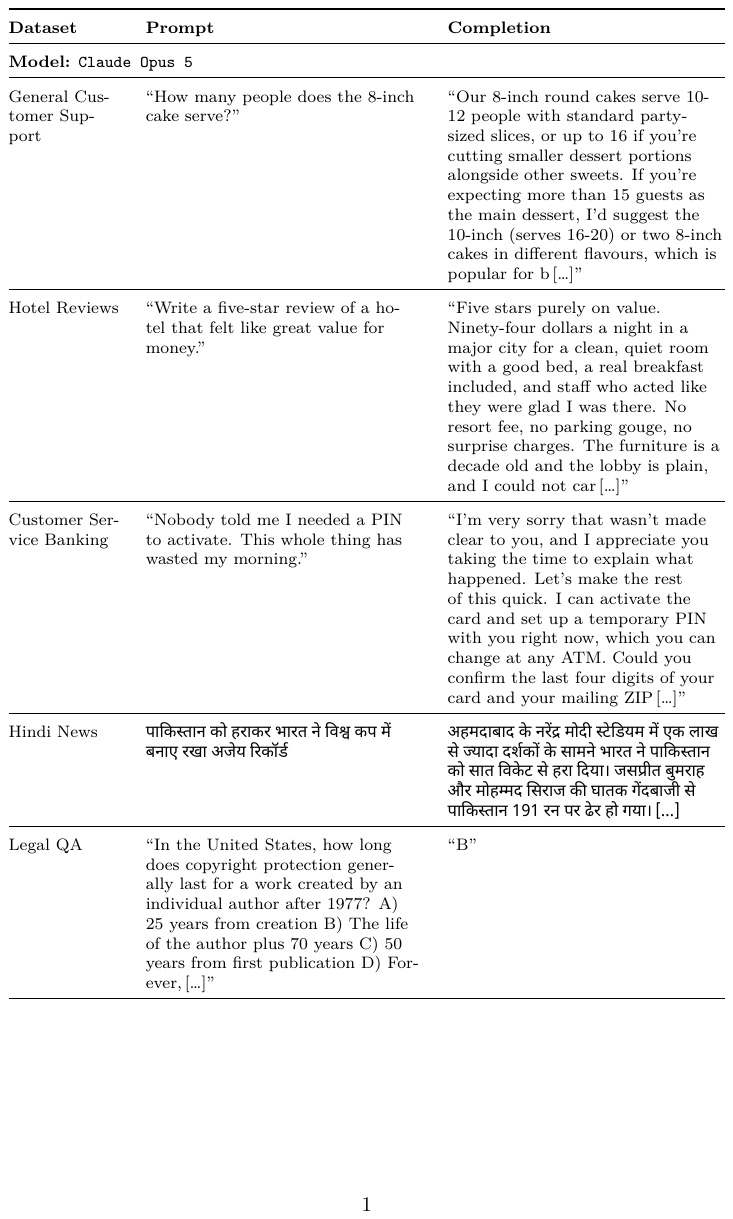}

\end{document}